\documentclass[letterpaper]{article} 
\usepackage{aaai25}  
\usepackage{times}  
\usepackage{helvet}  
\usepackage{courier}  
\usepackage[hyphens]{url}  
\usepackage{graphicx} 
\usepackage{natbib}  
\usepackage{caption} 
\usepackage{algorithm}
\usepackage{algorithmic}
\usepackage{amssymb}
\usepackage[table,xcdraw]{xcolor}
\usepackage{multirow}
\usepackage{newfloat}
\usepackage{listings}
\usepackage{amsmath}
\DeclareCaptionStyle{ruled}{labelfont=normalfont,labelsep=colon,strut=off} 
\floatstyle{ruled}
\newfloat{listing}{tb}{lst}{}
\floatname{listing}{Listing}
\title{Geodesic Flow Kernels for Semi-Supervised Learning on Mixed-Variable Tabular Dataset}
\author{
    Yoontae Hwang\textsuperscript{\rm 1}, Yongjae Lee\textsuperscript{† \rm 2}
}
\affiliations{
    \textsuperscript{\rm 1}University of Oxford\\
    \textsuperscript{\rm 2}Ulsan National Institute of Science and Technology(UNIST)\\
    \textsuperscript{\rm 1}yoontae.hwang@eng.ox.uk     \textsuperscript{\rm 2}yongjaelee@unist.ac.kr
}

\usepackage{bibentry}

\begin{document}

\maketitle

\begin{abstract}
Tabular data poses unique challenges due to its heterogeneous nature, combining both continuous and categorical variables. Existing approaches often struggle to effectively capture the underlying structure and relationships within such data. We propose GFTab (Geodesic Flow Kernels for Semi-Supervised Learning on Mixed-Variable Tabular Dataset), a semi-supervised framework specifically designed for tabular datasets. GFTab incorporates three key innovations: 1) Variable-specific corruption methods tailored to the distinct properties of continuous and categorical variables, 2) A Geodesic flow kernel based similarity measure to capture geometric changes between corrupted inputs, and 3) Tree-based embedding to leverage hierarchical relationships from available labeled data. To rigorously evaluate GFTab, we curate a comprehensive set of 21 tabular datasets spanning various domains, sizes, and variable compositions. Our experimental results show that GFTab outperforms existing ML/DL models across many of these datasets, particularly in settings with limited labeled data. \end{abstract}

%

\label{ch:ch1}
\begin{figure*}[ht]
\small{
  \centering
  \includegraphics[width=\linewidth]{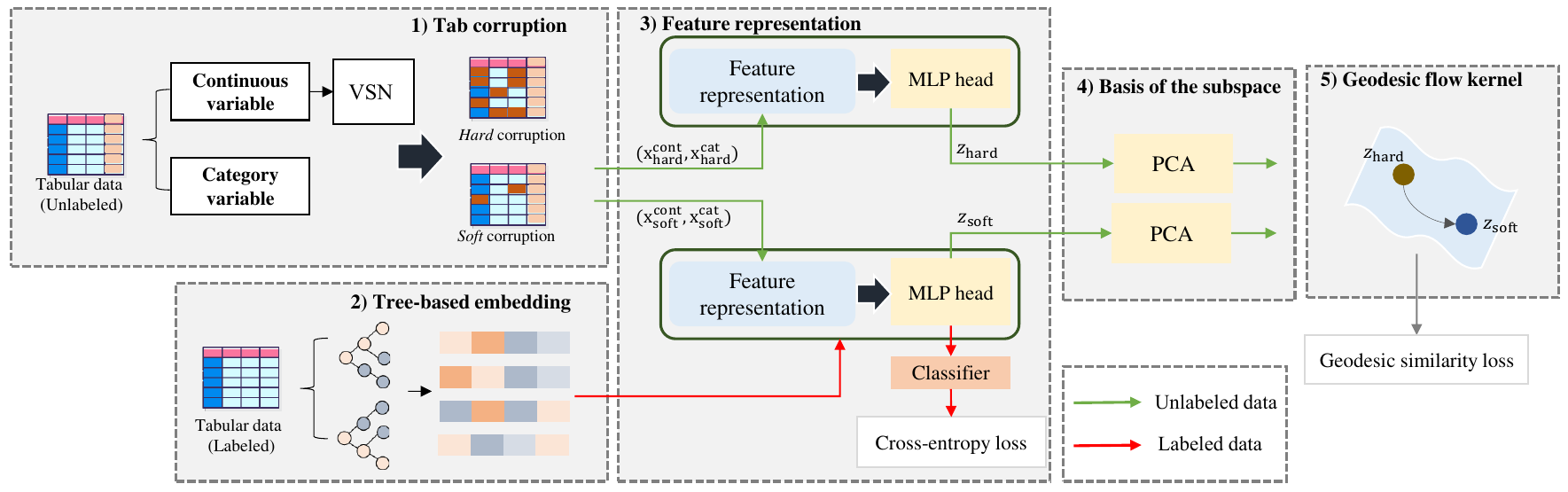}}
  \caption{The proposed model (GFTab) is a semi-supervised framework specifically designed for handling tabular data. }
  \label{figure_1}
 \vspace{-12pt}
\end{figure*}

\section{Introduction}
One of the long-standing goals in ML has been to discover structure and pattern in data. While these efforts have been very successful in handling images, audio, and text, there is still room for improvement in handling tabular data. There are \textit{three main reasons} why handling tabular data is difficult. One possible reason for this is that, unlike images and text, which share certain structural aspects within Euclidean space, tabular data do not appear to have such solid structural similarities. For instance, in an image, adjacent pixels are often observed to have similar color values and belong to the same object. However, in a tabular dataset, adjacent rows or columns may not necessarily have any inherent relationship or similarity. In most cases, the nature of tabular data would not change at all even if the order of rows and columns are shuffled.

Second, tabular datasets may present certain challenges due to the presence of both continuous and categorical variables, each with distinct characteristics. Continuous variables are characterized by an infinite number of potential values within a specified range, whereas categorical variables are typically represented by values from a finite set. Moreover, while continuous variables can be ordered, categorical variables cannot be ordered in general. It would be beneficial to consider the inherent differences between these variable types \cite{gorishniy2021revisiting}, rather than treating them identically as is often done in previous studies. Additionally, in real world, the proportion of continuous and categorical variables can vary considerably across different datasets, which can make it challenging to develop approaches that are widely applicable. This heterogeneity contributes to the domain-specific nature of tabular research and hinders the creation of universally applicable methods. It is possible that studies on tabular datasets have overlooked these complexities, evaluating performance only on benchmark data with predominantly continuous variables, which may not fully represent real-world scenarios.

Finally, real-world tabular datasets often lack labels, which are crucial for supervised learning tasks. In the absence of labels, discovering meaningful patterns and structures within the data becomes an even more daunting task. Without labels to guide the learning process, the model must rely on the intrinsic structure and relationships within the data to uncover relevant patterns, which can be particularly difficult given the lack of inherent structural similarities and the heterogeneous nature of tabular datasets. 

Some studies \cite{bahri2021scarf, ucar2021subtab, yoon2020vime} have demonstrated encouraging outcomes in addressing tabular datasets with limited labeled data. These methods employ corruption techniques to maximize the similarity between the original input and corrupted versions, thereby enabling the models to learn meaningful representations. However, these methods disregard variable types and have primarily been evaluated on datasets containing only continuous variables. Consequently, their effectiveness may be limited when dealing with real-world tabular datasets that often contain a mix of continuous and categorical variables \cite{Borisov_2022}. 

In this paper, we introduce GFTab: \textbf{G}eodesic \textbf{F}low Kernel on a Mixed-variable \textbf{Tab}ular dataset, an approach designed to address the unique challenges inherent in tabular data, particularly in semi-supervised learning settings. GFTab consists of three key components:

\begin{itemize}
    \item Variable-specific corruption methods that account for the distinct characteristics of continuous and categorical variables.
    \item  Geodesic flow kernel that captures the geometric relationships between corrupted representations on the Grassmann manifold.
    \item  Tree-based embedding technique that leverages labeled data to learn hierarchical feature relationships.
\end{itemize}

By integrating these components, GFTab is able to effectively learn robust representations of tabular data, even in scenarios with limited labeled samples or noisy labels. Our framework is designed to be applicable across a wide range of tabular datasets and domains. In the following sections, we describe each component of GFTab in detail, present our proposed benchmark datasets for evaluating tabular learning methods, and demonstrate GFTab's effectiveness through extensive experiments.

\label{ch:ch2}
\section{Related works}
{\bfseries Tabular DL Models.} Due to the importance of tabular data in real-world industries, ML/DL models for tabular data have been actively studied recently. The TabTransformer \cite{huang2020tabtransformer} generated embeddings on categorical variables to learn the context of tabular data in a robust way, and it performed well in terms of prediction Accuracy. However, they only used a simple MLP layer for continuous variables. AutoInt \cite{Song_2019} is designed to specifically learn the interactions between features in high-dimensional tabular data sets. The FT-transformer \cite{gorishniy2021revisiting} used Transformer layers, but it treated all features in the same way regardless of their variable types (categorical and continuous). Furthermore, it becomes computationally inefficient when there are a large number of features. NODE \cite{popov2020neural} proposed a method for training deep ensembles of oblivious differentiable decision trees on tabular data and demonstrated strong performance on only a few datasets. The TabGNN \cite{guo2021tabgnn} modeled the relationships within the tabular data samples using graph neural networks in an explicit and systematic manner. However, its numerical experiments are limited to large-sized tabular datasets only. NPTs \cite{kossen2021self} used self-attention to explicitly infer the relationships between data points like TabGNN. However, it is computationally expensive and has only been demonstrated on a small number of datasets. The GRANDE \cite{marton2024grande} model proposed an ensemble learning method to integrate multiple neural networks. TabPFN \cite{hollmann2023tabpfn} introduced a probabilistic approach for tabular data, leveraging Bayesian principles to provide robust uncertainty estimates and improve prediction accuracy. However, to the best of our knowledge, most of the models discussed above have been tested using datasets with 100\% labels, which is not representative of real world applications.

{\bfseries Tabular Self- and Semi-Supervised Learning} Self-supervised learning techniques have been actively studied in computer vision \cite{chen2020simple, jing2020self, zhai2019s4l} and natural language processing \cite{lee2019latent, qiu2020pre, ruder2018strong, song2020mpnet}, but not much research has been conducted for tabular data. The TABBIE \cite{iida2021tabbie}, TaBERT \cite{yin2020tabert} and TaPas \cite{herzig2020tapas} were developed to learn cell-wise, column-wise, and row-wise representations, respectively, in order to understand complex table semantics or numerical trends from PDF files or tables. The SAINT \cite{somepalli2021saint}, TabNet \cite{arik2021tabnet} and VIME \cite{yoon2020vime} were designed to be able to restore the original values of data samples after being injected with various types of noise. However, these works did not make a distinction between continuous and categorical variables. SCARF \cite{bahri2021scarf} was inspired by popular techniques used in computer vision, such as color distortion, random cropping, and blurring. It used these methods to corrupt a random subset of features and form a view for tabular data. SubTab \cite{ucar2021subtab} introduced a new framework that divides input features into multiple subsets and reconstructs the data from these subsets in an autoencoder setting. Similar to the other models, SCARF and SubTab also applied the same corruption methods regardless of variable type (continuous or categorical). In contrast to existing models, our model uses different corruption methods for continuous and categorical variables that are specifically designed to reflect the characteristics of each variable type. 
The VIME was the first to propose a semi-supervised learning framework for tabular data based on consistency regularization. Furthermore, to the best of our knowledge, there are no other semi-supervised learning studies for tabular data. Although it used different loss functions for continuous and categorical variables in the reconstruction and mask vector estimation, it still applied the same corruption method to both variable types as discussed before.


\label{ch:ch3}
\section{Methodology} \label{GFTab_section_3}
When learning from mixed-variable tabular dataset in a semi-supervised manner, a fundamental question arises:

\textit{How should we design inductive biases for tabular data?}

We propose GFTab (\textbf{G}eodesic \textbf{F}low Kernel on Mixed-variable \textbf{Tab}ular data) to exploit the unique characteristics of mixed-variable tabular datasets. Our framework includes: 1) Variable-specific corruption methods for continuous and categorical variables, 2) Geodesic flow kernel based similarity measure to capture geometric changes between corrupted inputs, and 3) Tree-based embedding to learn hierarchical relationships from labeled data.

\subsection{The GFTab Framework I : Corruption Process} \label{corruption_method}
\textbf{Semi-Supervised Setting} Suppose that there are $N_{l}$ labeled samples $D_l$ = $\{x_{i},y_{i}\}_{i=1}^{N_{l}}$  $\subseteq$ $\mathbb{R}^{M} \times \mathbb{R}$ and $N_{u}$ unlabeled samples $D_{u}=\{x_{i}\}_{i=1}^{N_u} \subseteq \mathbb{R}^{M}$, with $N_u$ significantly greater than $N_l$. In this scenario, $x_{i} \in \mathbb{R}^{M}$ represents an example, and $y_{i}\in \mathbb{R}$ is its corresponding label. We simplify by considering $y_{i}$ as a scalar, representing a class  $k \in \mathcal{K}$. For analytical reasons, we introduce $y_{i}^{k}$ as a binary indicator, which is set to 1 if $y_{i}=k$ and 0 otherwise. In this work, $x_i$ is broken down into continuous components $x_{i}^{\text{cont}} \in \mathbb{R}^{M_{\text{cont}}}$ and categorical components  $x_{i}^{\text{cat}} \in \mathbb{R}^{M_{\text{cat}}}$. For convenience, we will omit the subscript $i$ in the following discussions.

\textbf{Continuous Variable Corruption} For continuous variables, we firstly employ a variable selection network (VSN) to extract an effective feature representation. The VSN consists of two Gated Residual Networks (GRNs) \cite{lim2021temporal}: one for each element $x^{m, \text{cont}}$ of $x^{\text{cont}}$ (parameterized by $\theta_1$), and the other for the entire $x^{\text{cont}}$ (parameterized by $\theta_2$). The output of the first GRN is denoted as $\xi^{m}$=$\text{GRN}_{(\theta_1 )}(x^{m, \text{cont}})$, while the output of the second GRN is passed through a softmax function to obtain importance weights $v = \text{Softmax}(\text{GRN}_{(\theta_2)}(x^{\text{cont}}))$. The $m$-th element of $v$, denoted as $v^{m}$, serves as an importance weight for the corresponding continuous variable $x^{m, \text{cont}}$. See Appendix C for empirical validation of this feature selection capability. The final representation of continuous variables, $\Xi^{\text{cont}}$, is obtained by combining the weighted outputs of the first GRN:
\vspace{-0.2cm}
\begin{equation}
\Xi^{\text{cont}} = \sum_{m=1}^{M_{\text{cont}}} v^{m}\xi^{m}    
\label{eq:weight}
\end{equation}

To create corrupted views of continues variables, we generate soft and hard \textit{views}, $\text{x}_{\text{soft}}^{\text{cont}}$ and $\text{x}_{\text{hard}}^{\text{cont}}$ by randomly shuffling the row and columns within the $\Xi^{\text{cont}}$. Here, we define the permutation matrix $\textbf{P}$ of size $M_{\text{cont}} \times M_{\text{cont}}$.

\begin{equation}
\begin{aligned}
    \text{x}_{\text{soft}}^{\text{cont}} = \lambda \Xi^{\text{cont}} + (1-\lambda)\Xi^{\text{cont}}\textbf{P} \\
    \text{x}_{\text{hard}}^{\text{cont}}  = (1-\lambda) \Xi^{\text{cont}} + \lambda \Xi^{\text{cont}}\textbf{P}
\end{aligned}
\label{eq:corruption}
\end{equation}

In this case, the equation (\ref{eq:corruption}) generate corrupted \textit{views} of the continuous features $\Xi^{\text{cont}}$. The soft view $\text{x}_{\text{soft}}^{\text{cont}}$ mixes the original features with a randomly permuted version, while the hard view $\text{x}_{\text{hard}}^{\text{cont}}$ puts more emphasis on the permuted features. Thus, the soft views retains more of the original structure, while the hard views is more randomized.

\textbf{Categorical Variable Corruption} Unlike continuous variables,  Because ordered categories allow for the concept of relative closeness, unordered categories lack this inherent structure. categorical variables present unique challenges "\textit{how to corrupt categorical variable?}" due to their discrete nature and the distinction between ordered and unordered types. The motivation behind this and related analysis results can be found in Figure \ref{figure_cat}. 

Also, existing methods often overlook this distinction, leading to suboptimal results. For example, simply injecting noise \cite{bahri2021scarf, ucar2021subtab} or estimating masking \cite{yoon2020vime} may not sufficiently capture their complexity, especially when dealing with imbalanced distributions or variables with identical discrete values across different items. To address this, we propose a corruption technique tailored specifically to categorical variables. For \textit{categorical} variables, we determine the corrupted ratio of each variable. Let $k=[k^{1},...,k^{M_{\text{cat}}}]$ be the mask vector where $k^{m}$'s are independently sampled from over the set $[-\text{s}^{m}(\gamma), \text{s}^{m}(\gamma)]\backslash \{0\}$. Here, $\text{s}^{m}(\gamma)$ is size of the neighborhood for $k^{m}$ with corruption rate $\gamma$. Sampling from the set $[-1,1] \backslash$\{0\} would result in the most corruption. We then create soft and hard \textit{views} for categorical variables as follows:
\vspace{-0.1cm}
\begin{equation}
\begin{aligned}
    \text{x}_{\text{soft}}^{\text{cat}} &=  x^{\text{cat}} + \text{nh}_{\text{s}(\gamma)}(k) \\
    \text{x}_{\text{hard}}^{\text{cat}} &= x^{\text{cat}}  + \text{nh}_{\text{s}=1}(k)
\label{eq:cat_corruption}
\end{aligned}
\end{equation}
\vspace{-0.15cm}

Where the neighborhood function \( \text{nh}_{s(\gamma)} \) is defined as a random function that shifts each category \( c \) to \( c' \) within a neighborhood of size \( s(\gamma) \). The size \( s(\gamma) \) is determined such that the probability of any category shift from \( c \) to \( c' \) matches the corruption rate \( \gamma \). The following remark outlines how the neighborhood size is determined when the corruption rate is set arbitrarily. If the shifted value \( c' \) falls outside the range $[0, n]$, it is reset to \( c \). 

\textbf{Remark:} Consider a categorical variable with \( n \ge 2 \) categories. To achieve a corruption rate of at least \( \gamma \), the minimum size \( s \) of the neighborhood \( \text{nh}_{\text{s}(\gamma)} \) must satisfy the inequality $\text{s} \geq \left\lceil 2n(1 - \gamma) - 1 \right\rceil$.

\textbf{Feature Representation Module.} After creating separate views for continuous and categorical variables, the feature representation module aims to capture the relationships between all variables to learn a more comprehensive representation of the data. To achieve this, we combines the information from the soft and hard views of the data samples, denoted as $\text{x}_{\text{soft}}=$ ($\text{x}_{\text{soft}}^{\text{cont}}$, $\text{x}_{\text{soft}}^{\text{cat}}$) and $\text{x}_{\text{hard}}=$ ($\text{x}_{\text{hard}}^{\text{cont}}$, $\text{x}_{\text{hard}}^{\text{cat}}$), respectively. We apply the Transformer layers to the column embeddings $\textbf{e}(x_{\text{soft}}) \in \mathbb{R}^{\text{d}_{\text{emb}}}$ and $\textbf{e}(x_{\text{hard}}) \in \mathbb{R}^{\text{d}_{\text{emb}}}$.  First, a self-attention \cite{NIPS2017_3f5ee243} aggregates column embeddings with normalized importance:
\vspace{-0.1cm}
\begin{equation}
    \vspace{-0.1cm}
    \text{Attention}(\text{Q},\text{K},\text{V}) = \text{Softmax}(\frac{\text{QK}^{\top}}{\text{d}_{{\text{attn}}}})\text{V}
\end{equation}
Here, $\text{Q}=\text{W}_{Q}\textbf{e} \in \mathbb{R}^{\text{d}_{\text{emb}}}_{\text{attn}}$, $\text{K}=\text{W}_{\text{K}}\textbf{e} \in \mathbb{R}^{\text{d}_{\text{emb}}}_{\text{attn}}$ and $\text{V}=\text{W}_{\text{V}}\textbf{e} \in \mathbb{R}^{\text{d}_{\text{emb}}}_{\text{v}}$ are queries, keys, and values, respectively. \textbf{e} is the collection of column embeddings, $\text{W}_\text{Q}$, $\text{W}_\text{K}$ and $\text{W}_\text{V}$ are learnable weight matrices. In actual implementation, multi-head attention is used to further improve the feature extraction. The output is then transformed back into an embedding of dimension $\text{d}_{\text{lin}}$ through a fully connected layer. Then, the representation embedding $z_{\text{hard}} \in \mathbb{R}^{d_{\text{lin}}}$ and $z_{\text{soft}} \in \mathbb{R}^{d_{\text{lin}}} $ are obtained after passing through feed-forward layers. 

\subsection{The GFTab Framework II: Geodesic Flow Kernel} \label{geodesic-kernel-flow}
The motivation for using a geodesic flow kernel in mixed-variable tabular data lies in the inherent challenges of measuring distances. In the context of tabular data, the distance between two points cannot always be easily or meaningfully quantified using conventional distance metrics such as InfoNCE \cite{oord2018representation}, Barlow \cite{barlow} and so on. 

For example, a small change in a continuous variable might not significantly affect the overall relationship between data points. However, a single change in a categorical variable could result in a completely different context or classification. This variability necessitates a more sophisticated approach to capturing the relationships between data points. The related analysis results can be found in Figure \ref{figure_loss}. 

In this section, we address this by modeling geodesic flow kernel. 
For this, we leverage the assumption that data lie in a low-dimensional linear subspace to derive features from both $z_{\text{soft}}$ and $z_{\text{hard}}$, modeling them with a basis $\textbf{P} \in \mathbb{R}^{d_{\text{lin}} \times D}$.


\textbf{DEF 1 (Grassmannian)} Grassmannian $\mathbf{G}(D, d_{\text{lin}})$, which is the collection of all $D$-dimensional linear subspaces $\mathbb{R}^{d_{\text{lin}}\times D}$, is a smooth Riemannian manifold. Also, an element $\mathbf{P}$ of $\mathbf{G}(D, d_{\text{lin}})$ can be specified by a basis, i.e., $d_{\text{lin}} \times D$ matrix with orthogonal columns.


In the context of the GFTab, the concept of Grassmannians facilitates a robust approach to feature representation from a tabular dataset by allowing us to model the space of subspaces formed by features.  By representing different corrupted \textit{views}, such as $z_{\text{soft}}$ and $z_{\text{hard}}$, as points on the Grassmannian, we gain the ability to learn their geometric relationship.  This approach aids in identifying invariant features that are crucial for learning under corruption intensity. 

To capture this relationship, we introduce the concept of geodesic flow \cite{gallivan2003efficient}, denoted as $\textbf{GF}(\pi)$. Let $\textbf{P}_{\text{hard}}$ and $\textbf{P}_{\text{soft}}$ denote the sets of basis vector for the subspaces corresponding to the hard and soft representations, respectively. The geodesic flow between $\textbf{P}_{\text{hard}}$ and $\textbf{P}_{\text{soft}}$, denotedy by \textbf{GF}, is parameterized as $\textbf{GF}: \pi\in[0,1] \rightarrow  \textbf{GF}(\pi) \in \textbf{G}(D,d_{\text{lin}})$ such that $\textbf{GF}(0) = \textbf{P}_{\text{hard}}$ and $\textbf{GF}(1) = \textbf{P}_{\text{soft}}$. For $\pi \in (0,1)$, 
\vspace{-0.1cm}
\begin{equation}
\begin{aligned}
\vspace{-0.1cm}
\mathbf{GF}(\pi) &= \left[\textbf{P}_{\text{hard}}  \quad \textbf{R}_{\text{hard}}\right]\left[\begin{array}{ll}
\textbf{U}_1 \boldsymbol{\Gamma}(\pi) & 0 \\
0 & -\textbf{U}_2 \boldsymbol{\Sigma}(\pi)
\end{array}\right] \\
&= \textbf{P}_{\text{hard}} \textbf{U}_1 \boldsymbol{\Gamma}(\pi)-\textbf{R}_{\text{hard}} \textbf{U}_2 \boldsymbol{\Sigma}(\pi)
\end{aligned}
\end{equation}
Here, the $\textbf{R}_{\text{hard}} \in \mathbb{R}^{d_{\text{lin}} \times\left(d_{\text{lin}}-D\right)}$ is the orthogonal complement of $\textbf{P}_{\text{hard}}$. 

According to the gSVD \cite{van1976generalizing}, the matrices $\textbf{P}^{\top}_{\text{hard}} \textbf{P}_{\text{soft}}$ and $\textbf{R}^{\top}_{\text{hard}} \textbf{P}_{\text{soft}}$ share identical right singular vectors, $\textbf{V}$. Using gSVD, we can decompose as here:
\vspace{-0.1cm}
\begin{equation}
\begin{aligned}
\textbf{P}^{\top}_{\text{hard}}\textbf{P}_{\text{soft}} = \textbf{U}_{1}\boldsymbol{\Gamma}\textbf{V}^{\top} \\
\textbf{R}^{\top}_{\text{hard}}\textbf{P}_{\text{soft}} = \textbf{U}_{2}\boldsymbol{\Sigma}\textbf{V}^{\top} 
\end{aligned}
\end{equation}

\noindent where $\boldsymbol{\Gamma}$ and $\boldsymbol{\Sigma}$ have diagonal elements $\text{cos}(\theta_{i})$ and $\text{sin}(\theta_{i})$ for $i=1,...,D$. the $\theta_{i}$ are the principal angle \cite{mandolesi2019grassmann} between $\textbf{P}_{\text{hard}}$ and $\textbf{P}_{\text{soft}}$. 

By varying the parameter $\pi$ from 0 to 1, we generate intermediate subspaces that capture the gradual transition from the $z_{\text{hard}}$ to the $z_{\text{soft}}$. This transition is facilitated by the geodesic flow kernel, which smoothly varies $z_{\text{hard}}$ along the geodesic path to $z_{\text{soft}}$. The purpose of this smooth variation is to learn the underlying relationships in the tabular data, allowing us to obtain a representation that is robust to real-world noise. To quantify this transition and capture these noise-resistant relationships, we project the encoded representations onto the subspace $\textbf{GF}(\pi)$ at each point along the geodesic. To formalize this process and provide a measure of similarity between the encoded representations, we introduce the geodesic flow kernel.

\textbf{DEF 2. (Geodesic Flow Kernel)} Let $z_{\text{hard}}$ and $z_{\text{soft}}$ are the encoded representation, respectively. The, the geodesic flow kernel is defined as:
\vspace{-0.1cm}
\begin{equation}
    \vspace{-0.1cm}
    z_{\text{hard}}^{\top}\textbf{A}z_{\text{soft}} = \int_0^1\left(\mathbf{G F}(\pi)^{\top} z_{\text{hard}}\right)^{\top}\left(\mathbf{G F}(\pi)^{\top} z_{\text{soft}}\right) d \pi
\end{equation}
where $\textbf{A} \in \mathbb{R}^{D \times D}$ is positive semi-definite matrix. 

Fortunately, we do not necessary to compute all projection into the subspace. By \cite{gong2012geodesic, simon2021learning}, we can get matrix \textbf{A} in a closed form:
\vspace{-0.1cm}
\begin{equation}
\textbf{A}=\left[\begin{array}{lll}
\textbf{P}_{\text{hard}} \textbf{U}_1 & \textbf{R}_{\text{soft}} \textbf{U}_2
\end{array}\right]\left[\begin{array}{ll}
\textbf{D} & \textbf{E} \\
\textbf{E} & \textbf{G}
\end{array}\right]\left[\begin{array}{l}
\textbf{U}_1^{\top} \textbf{P}_{\text{hard}}^{\top} \\
\textbf{U}_2^{\top} \textbf{R}_{\text{soft}}^{\top}
\end{array}\right]
\end{equation}

\noindent where $d_{1 i}=1+\frac{\sin \left(2 \theta_i\right)}{2 \theta_i}, e_{2 i}=\frac{\cos \left(2 \theta_i\right)-1}{2 \theta_i}$ and $g_{3 i}=1-\frac{\sin \left(2 \theta_i\right)}{2 \theta_i}$, which are the $i$-th elements of the diagonal matrices $\textbf{D}, \textbf{E}$, and $\textbf{G}$ respectively.

The matrix $\textbf{A} \in \mathbb{R}^{D \times D}$ captures the geometric relationship between $z_{\text{soft}}$ and $z_{\text{hard}}$ subspaces on $\mathbf{G}(D, d_{\text{lin}})$. In other words, by learn the variation between $z_{\text{hard}}$ and $z_{\text{soft}}$, our model extracts meaningful patterns, even with noise values common in real-world tabular datasets.
The geodesic similarity loss function $\mathcal{L}_{\text{sim}}$ in Equation (\ref{eq:GFTab_loss}) encourages the model to minimize the dissimilarity between the $z_{\text{hard}}$ and $z_{\text{soft}}$ representations, taking into account their geometric relationship on $\mathbf{G}(D, d_{\text{lin}})$. 
\vspace{-0.2cm}
\begin{equation}
    \vspace{-0.1cm}
    \mathcal{L}_{\text{sim}} =1-\frac{z_{\text{soft}}^{\top} \textbf{A} z_{\text{hard}}}{\left\|\sqrt{\textbf{A}}z_{\text{soft}}\right\|\left\|\sqrt{\textbf{A}} z_{\text{hard}}\right\|}
\label{eq:GFTab_loss}
\end{equation}

\begin{table*}[!htbp]
    \centering
{\fontsize{6.3}{6.8}\selectfont
\begin{tabular}{lccccccc}
\hline
\multicolumn{8}{l}{\textbf{Panel A. Datasets with more categorical variables}} \\
\multicolumn{1}{c}{\textbf{model}} &
  \textbf{Diabetes} &
  \textbf{Insurance} &
  \textbf{adult} &
  \textbf{bank} &
  \textbf{cmc} &
  \textbf{credit-approval} &
  \textbf{credit-g} \\ \hline
GFTab &
  {\color[HTML]{FE0000} 0.3826 \tiny$\pm$0.0139} &
  {\color[HTML]{FE0000} 0.4500 \tiny$\pm$0.0022} &
  {\color[HTML]{FE0000} 0.8023 \tiny$\pm$0.0070} &
  {\color[HTML]{FE0000} 0.7336 \tiny$\pm$0.0040} &
  \textbf{0.4625 \tiny{$\pm$0.0215}} &
  0.6738 \tiny{$\pm$0.0019} &
  {\color[HTML]{FE0000} 0.7433 \tiny$\pm$0.0424} \\
GRANDE &
  {\color[HTML]{3531FF} 0.3635 \tiny$\pm$0.0012} &
  0.4225 \tiny{$\pm$0.0001} &
  {\color[HTML]{3531FF} 0.7596 \tiny$\pm$0.0033} &
  {\color[HTML]{3531FF} 0.7017 \tiny$\pm$0.0042} &
  0.4306 \tiny{$\pm$0.0229} &
  {\color[HTML]{3531FF} 0.8521 \tiny$\pm$0.0008} &
  {\color[HTML]{3531FF} 0.6570 \tiny$\pm$0.0224} \\
TabPFN &
  0.2559 \tiny{$\pm$0.0089} &
  0.4312 \tiny{$\pm$0.0001} &
  \textbf{0.7530 \tiny{$\pm$0.0024}} &
  0.6102 \tiny{$\pm$0.0047} &
  0.4566 \tiny{$\pm$0.0027} &
  {\color[HTML]{FE0000} 0.8975 \tiny$\pm$0.0091} &
  0.4802 \tiny{$\pm$0.0409} \\
SCARF &
  0.2335 \tiny{$\pm$0.0001} &
  0.4312 \tiny{$\pm$0.0002} &
  0.4157 \tiny{$\pm$0.0244} &
  0.4699 \tiny{$\pm$0.0009} &
  0.3479 \tiny{$\pm$0.0425} &
  0.6993 \tiny{$\pm$0.0518} &
  0.4451 \tiny{$\pm$0.0377} \\
SubTab &
  0.2569 \tiny{$\pm$0.0187} &
  0.4312 \tiny{$\pm$0.0001} &
  0.7506 \tiny{$\pm$0.0035} &
  0.6626 \tiny{$\pm$0.0049} &
  {\color[HTML]{FE0000} 0.5012 \tiny$\pm$0.0141} &
  0.6728 \tiny{$\pm$0.0202} &
  0.6474 \tiny{$\pm$0.0388} \\
VIME &
  0.2611 \tiny{$\pm$0.0313} &
  {\color[HTML]{3531FF} 0.4314 \tiny{$\pm$0.0001}} &
  0.7369 \tiny{$\pm$0.0057} &
  \textbf{0.6932 \tiny{$\pm$0.0179}} &
  {\color[HTML]{3531FF} 0.4942 \tiny$\pm$0.0336} &
  0.7177 \tiny{$\pm$0.1637} &
  0.5397 \tiny{$\pm$0.0891} \\
XGBoost &
  \textbf{0.3534 \tiny{$\pm$0.0002}} &
  0.4312 \tiny{$\pm$0.0003} &
  0.7344 \tiny{$\pm$0.0006} &
  0.5246 \tiny{$\pm$0.0024} &
  0.4540 \tiny{$\pm$0.0213} &
  0.8276 \tiny{$\pm$0.0096} &
  \textbf{0.5968 \tiny{$\pm$0.0171}} \\
CatBoost &
  0.3552 \tiny{$\pm$0.0003} &
  0.4312 \tiny{$\pm$0.0001} &
  0.7428 \tiny{$\pm$0.0006} &
  0.4844 \tiny{$\pm$0.0103} &
  0.3694 \tiny{$\pm$0.0430} &
  {\color[HTML]{29261B} \textbf{0.8469 \tiny{$\pm$0.0082}}} &
  0.5590 \tiny{$\pm$0.0310} \\ \hline
\multicolumn{1}{c}{\textbf{model}} &
  \textbf{dresses-sales} &
  \textbf{fars} &
  \textbf{jasmine} &
  \textbf{kick} &
  \textbf{okcupid-stem} &
  \textbf{online-shoppers} &
  \textbf{shrutime} \\ \hline
GFTab &
  0.4165 \tiny{$\pm$0.0057} &
  {\color[HTML]{FE0000} 0.6199 \tiny$\pm$0.0104} &
  \textbf{0.7775 \tiny{$\pm$0.0041}} &
  {\color[HTML]{FE0000} 0.4928 \tiny$\pm$0.0044} &
  {\color[HTML]{3531FF} 0.4219 \tiny$\pm$0.0195} &
  \textbf{0.7871 \tiny{$\pm$0.0106}} &
  \textbf{0.7088 \tiny$\pm$0.0025} \\
GRANDE &
  0.4611 \tiny{$\pm$0.0256} &
  \textbf{0.5600 \tiny{$\pm$0.0122}} &
  0.7630 \tiny{$\pm$0.0044} &
  \textbf{0.4654 \tiny{$\pm$0.0012}} &
  {\color[HTML]{FE0000} 0.4412 \tiny$\pm$0.0117} &
  0.7405 \tiny{$\pm$0.0042} &
  {\color[HTML]{FE0000} 0.7355 \tiny$\pm$0.0034} \\
TabPFN &
  0.3671 \tiny{$\pm$0.0010} &
  0.5247 \tiny{$\pm$0.0104} &
  0.7491 \tiny{$\pm$0.0036} &
  0.4749 \tiny{$\pm$0.0010} &
  0.3889 \tiny{$\pm$0.0019} &
  {\color[HTML]{3531FF} 0.7954 \tiny$\pm$0.0012} &
  {\color[HTML]{3531FF} 0.7130 \tiny{$\pm$0.0069}} \\
SCARF &
  0.4694 \tiny{$\pm$0.0838} &
  0.1314 \tiny{$\pm$0.0095} &
  0.5839 \tiny{$\pm$0.0138} &
  0.4753 \tiny{$\pm$0.0010} &
  0.2787 \tiny{$\pm$0.0000} &
  0.4579 \tiny{$\pm$0.0003} &
  0.4459 \tiny{$\pm$0.0044} \\
SubTab &
  {\color[HTML]{FE0000} 0.5300 \tiny$\pm$0.0348} &
  0.5020 \tiny{$\pm$0.0035} &
  0.7078 \tiny{$\pm$0.0063} &
  {\color[HTML]{3531FF} 0.4840 \tiny{$\pm$0.0030}} &
  \textbf{0.4021 \tiny{$\pm$0.0118}} &
  0.6624 \tiny{$\pm$0.0096} &
  0.6808 \tiny{$\pm$0.0026} \\
VIME &
  {\color[HTML]{3531FF} 0.5069 \tiny{$\pm$0.0346}} &
  {\color[HTML]{3531FF} 0.5900 \tiny$\pm$0.0041} &
  0.7548 \tiny{$\pm$0.0050} &
  0.4758 \tiny{$\pm$0.0006} &
  0.3266 \tiny{$\pm$0.0180} &
  0.4583 \tiny{$\pm$0.0015} &
  0.5323 \tiny{$\pm$0.0421} \\
XGBoost &
  \textbf{0.5041 \tiny{$\pm$0.0553}} &
  0.4196 \tiny{$\pm$0.0002} &
  {\color[HTML]{3531FF} 0.7851 \tiny$\pm$0.0061} &
  0.4749 \tiny{$\pm$0.0002} &
  0.3546 \tiny{$\pm$0.0043} &
  {\color[HTML]{FE0000} 0.7964 \tiny$\pm$0.0046} &
  0.6863 \tiny{$\pm$0.0002} \\
CatBoost &
  0.4056 \tiny{$\pm$0.0804} &
  0.3571 \tiny{$\pm$0.0002} &
  {\color[HTML]{FE0000} 0.7940 \tiny$\pm$0.0029} &
  0.4749 \tiny{$\pm$0.0001} &
  0.3062 \tiny{$\pm$0.0059} &
  0.7847 \tiny{$\pm$0.0065} &
  0.6215 \tiny{$\pm$0.0107} \\ \hline
\multicolumn{8}{l}{} \\ \hline
\multicolumn{8}{l}{\textbf{Panel B. Datasets with more continouse variables}} \\
\multicolumn{1}{c}{\textbf{model}} &
  \textbf{KDD} &
  \textbf{Shipping} &
  \textbf{churn} &
  \textbf{eye-movements} &
  \textbf{nomao} &
  \textbf{qsar} &
  \textbf{road-safety} \\ \hline
GFTab &
  \textbf{0.7998 \tiny{$\pm$0.0070}} &
  {\color[HTML]{FE0000} 0.6493 \tiny$\pm$0.0020} &
  {\color[HTML]{FE0000} 0.7865 \tiny$\pm$0.0365} &
  0.5534 \tiny$\pm$0.0205 &
  {\color[HTML]{FE0000} 0.9411 \tiny$\pm$0.0030} &
  0.7867 \tiny{$\pm$0.0140} &
  {\color[HTML]{3531FF} 0.7553 \tiny$\pm$0.0042} \\
GRANDE &
  0.7848 \tiny{$\pm$0.0069} &
  0.6186 \tiny{$\pm$0.0045} &
  {\color[HTML]{3531FF} 0.7783 \tiny$\pm$0.0247} &
  0.5676 \tiny{$\pm$0.0102} &
  0.9113 \tiny{$\pm$0.0067} &
  0.7729 \tiny{$\pm$0.0110} &
  {\color[HTML]{FE0000} 0.7566 \tiny$\pm$0.0017} \\
TabPFN &
  0.7722 \tiny{$\pm$0.0045} &
  {\color[HTML]{3531FF} 0.6432 \tiny$\pm$0.0024} &
  \textbf{0.7657 \tiny{$\pm$0.0174}} &
  {\color[HTML]{FE0000} 0.5854 \tiny$\pm$0.0020} &
  0.8893 \tiny{$\pm$0.0032} &
  \textbf{0.8254 \tiny{$\pm$0.0000}} &
  0.7389 \tiny{$\pm$0.0015} \\
SCARF &
  0.5603 \tiny{$\pm$0.0183} &
  0.6213 \tiny{$\pm$0.0141} &
  0.4624 \tiny{$\pm$0.0005} &
  0.4878 \tiny{$\pm$0.0167} &
  0.5065 \tiny{$\pm$0.0217} &
  0.6126 \tiny{$\pm$0.0528} &
  0.4976 \tiny{$\pm$0.0023} \\
SubTab &
  0.6634 \tiny{$\pm$0.0269} &
  0.5514 \tiny{$\pm$0.0091} &
  0.7539 \tiny{$\pm$0.014} &
  {\color[HTML]{3531FF} 0.5711 \tiny{$\pm$0.0060}} &
  \textbf{0.9290 \tiny{$\pm$0.0038}} &
  {\color[HTML]{3531FF} 0.8404 \tiny$\pm$0.0139} &
  0.6750 \tiny{$\pm$0.0006} \\
VIME &
  0.7042 \tiny{$\pm$0.0164} &
  0.6358 \tiny{$\pm$0.0148} &
  0.7051 \tiny{$\pm$0.0304} &
  0.5524 \tiny{$\pm$0.0277} &
  {\color[HTML]{3531FF} 0.9361 \tiny$\pm$0.0028} &
  {\color[HTML]{FE0000} 0.8538 \tiny$\pm$0.0158} &
  0.7528 \tiny{$\pm$0.0025} \\
XGBoost &
  {\color[HTML]{3531FF} 0.8001 \tiny$\pm$0.0081} &
  0.6247 \tiny{$\pm$0.0081} &
  0.5231 \tiny{$\pm$0.0001} &
  0.5533 \tiny{$\pm$0.0187} &
  0.9078 \tiny{$\pm$0.0002} &
  0.8001 \tiny{$\pm$0.0035} &
  0.7452 \tiny{$\pm$0.0250} \\
CatBoost &
  {\color[HTML]{FE0000} 0.8138 \tiny$\pm$0.0073} &
  \textbf{0.6416 \tiny{$\pm$0.0038}} &
  0.5163 \tiny{$\pm$0.0413} &
  \textbf{0.5685 \tiny{$\pm$0.0161}} &
  0.8936 \tiny{$\pm$0.0040} &
  0.7824 \tiny{$\pm$0.0252} &
  \textbf{0.7545 \tiny{$\pm$0.0153}} \\ \hline
\end{tabular}}
    \caption{Comparison of F1 score between GFTab and baseline models on 21 tabular benchmark datasets in 20\% labeled training setting. The best performing method is highlighted in \textcolor{red}{red} and the second best in \textcolor{blue}{blue}, while the third best is \textbf{bold}.}
    \label{f1_02}
    \vspace{-0.5cm}
\end{table*}

\subsection{The GFTab Framework III: Tree-Based Embedding} \label{tree-based-embedding} For tabular data, the relationships between columns are crucial, as they can significantly impact the overall data structure and interpretation. However, most existing tabular DL models fail to capture comprehensive relationships between many different columns. To address this limitation, we employ a tree-based embedding approach inspired by DATE \cite{kim2020date} to effectively leverage labeled data. Note that tree-based models are known to be effective in capturing various relationships among different columns. A GBDT $\mathcal{T} = \{\mathcal{V}, \mathcal{E}\}$ is used, where $\mathcal{V} = \{v_R\} \cup \mathcal{V}_I \cup \mathcal{V}_L$ represents nodes (root, internal, and leaf) and $\mathcal{E}$ is set of edges.
Each leaf node $v_L \in \mathcal{V}_L$ is assigned a learnable embedding vector $\textbf{s}_L \in \mathbb{R}^{\text{d}_{\text{emb}}}$. For data point $x_i$, activated leaves are represented by binary vector $\textbf{p}_i$. The tree-based embedding $\textbf{S}_i \in \mathbb{R}^{\text{L} \times \text{d}_{\text{emb}}}$ is:
\begin{equation}
\textbf{S}_i = \phi(\textbf{p}_{i,1}\textbf{s}_1, \textbf{p}_{i,2}\textbf{s}_2, \ldots, \textbf{p}_{i,|\mathcal{V}_L|}\textbf{s}_{|\mathcal{V}_L|}),
\end{equation}
where $\phi(\cdot)$ removes all-zero rows. The tree embeddings are then processed through the feature representation module and a feed-forward layer. See Appendix D for empirical validation of this embeddings capability. So, GFTab's overall loss combines geodesic similarity loss $\mathcal{L}_{\text{sim}}$ and supervised cross-entropy loss $\mathcal{L}_{\text{ce}}$:
\begin{equation}
\mathcal{L}_{\text{GFTab}} = \mathcal{L}_{\text{sim}} + \beta \mathcal{L}_{\text{ce}},
\end{equation}

\noindent where $\mathcal{L}_{\text{ce}}$ is cross-entropy loss for labeled data and $\beta$ is a balance parameter that controls the relative importance of the supervised loss.

\section{Tabular Benchmarks Dataset} \label{section_4}
We configured benchmark datasets to properly evaluate the performance of our proposed method. We selected 21 datasets after carefully reviewing more than 4,000 datasets including OpenML (3,953 datasets), AMLB \cite{amlb} (71 datasets), and \cite{grinsztajn2022tree}(22 datasets). Detailed criteria are given below.

\begin{itemize}
\item {\bfseries Preprocessing.} Datasets with more than 30\% missing values were excluded. For the remaining datasets, columns with more than 30\% missing values were removed. Also, redundant categorical variables, which have only one category, were removed.

\item {\bfseries Variable types.} In order to evaluate tabular models in a more real-world like environment, we selected datasets with both continuous and categorical variables. Surprisingly, around 60\% of the entire datasets did not satisfy this condition.  

\item {\bfseries Data distribution.} We assume that data samples are i.i.d. Hence, datasets with certain distributional structure (sequential or temporal) were excluded. Also, we eliminated datasets with too simple distributions, which can be easily predicted with high Accuracy by naive models. Artificially generated datasets were excluded as well. Lastly, as this study focuses on classification tasks, datasets for regression tasks were not considered.

\item {\bfseries Dataset size.} size Most previous studies did not evaluate their models with datasets of different sizes. For more comprehensive evaluation, we selected datasets with different sizes: small-sized ($\sim$10,000 samples), medium-sized (10,000$\sim$50,000), and large-sized (50,000$\sim$). 
\end{itemize}
More details on dataset can be found in Appendix A. 


\label{ch:ch4}
\section{Experiment}
We evaluate GFTab on a diverse set of tabular datasets to assess its performance and robustness across different domains and data characteristics. It is important to note that the performance of machine learning models on tabular data often depends on the specific dataset and its inherent properties. Therefore, we include a wide range of datasets in our experiments to account for this variability and to provide a fair and thorough evaluation of GFTab. 
Our code will be publicly available on Github\footnote{https://github.com/Yoontae6719/Geodesic-Flow-Kernels-for-Semi-Supervised-Learning-on-Mixed-Variable-Tabular-Dataset}.

\subsection{Implementation details} We present the implementation details of evaluation metrics and baseline models. For other settings such as hyperparameters and experiment settings, see Appendix B.

\begin{table*}[!htbp]
\centering
    \centering
{\fontsize{6.3}{6.8}\selectfont
\begin{tabular}{lccccccc}
\hline
\multicolumn{8}{l}{\textbf{Panel A. Datasets with more categorical variables}} \\
\multicolumn{1}{c}{\textbf{model}} &
  \textbf{Diabetes} &
  \textbf{Insurance} &
  \textbf{adult} &
  \textbf{bank} &
  \textbf{cmc} &
  \textbf{credit-approval} &
  \textbf{credit-g} \\ \hline
GFTab  &
  {\color[HTML]{FE0000} 0.3879 \tiny$\pm$0.0013} &
  {\color[HTML]{FF0000} 0.4765 \tiny$\pm$0.0022} &
  {\color[HTML]{FE0000} 0.7680 \tiny$\pm$0.0015} &
  {\color[HTML]{FE0000} 0.6972 \tiny$\pm$0.0040} &
  {\color[HTML]{3531FF} 0.4598 \tiny$\pm$0.0284} &
  {\color[HTML]{FE0000} 0.7853 \tiny{$\pm$0.0054}} &
  \textbf{0.5376 \tiny$\pm$0.0144} \\
GRANDE &
  {\color[HTML]{3531FF} 0.3730 \tiny$\pm$0.0046} &
  {\color[HTML]{3531FF} 0.4594 \tiny{$\pm$0.0034}} &
  {\color[HTML]{3531FF} 0.7329 \tiny$\pm$0.0098} &
  {\color[HTML]{3531FF} 0.6772 \tiny$\pm$0.0073} &
  0.4316 \tiny{$\pm$0.0337} &
  0.7330 \tiny{$\pm$0.0228} &
  0.4958 \tiny{$\pm$0.0359} \\
TabPFN &
  0.2335 \tiny{$\pm$0.0120} &
  0.4312 \tiny{$\pm$0.0000} &
  0.7034 \tiny{$\pm$0.0029} &
  0.4689 \tiny{$\pm$0.0012} &
  {\color[HTML]{FE0000} 0.4868 \tiny$\pm$0.0070} &
  0.7603 \tiny{$\pm$0.0010} &
  0.4118 \tiny{$\pm$0.0100} \\
SCARF &
  0.2335 \tiny{$\pm$0.0010} &
  0.4312 \tiny{$\pm$0.0000} &
  0.4222 \tiny{$\pm$0.0116} &
  0.4710 \tiny{$\pm$0.0035} &
  0.3010 \tiny{$\pm$0.0310} &
  0.6312 \tiny{$\pm$0.0854} &
  0.4372 \tiny{$\pm$0.0374} \\
SubTab &
  0.2450 \tiny{$\pm$0.0021} &
  \textbf{0.4328 \tiny{$\pm$0.0011}} &
  0.7325 \tiny{$\pm$0.0039} &
  0.6136 \tiny{$\pm$0.0059} &
  \textbf{0.4551 \tiny{$\pm$0.0143}} &
  0.3796 \tiny{$\pm$0.0237} &
  {\color[HTML]{FF0000} 0.6007 \tiny{$\pm$0.0144}} \\
VIME &
  0.2219 \tiny{$\pm$0.0097} &
  0.4101 \tiny{$\pm$0.0001} &
  0.7142 \tiny{$\pm$0.0082} &
  \textbf{0.6755 \tiny{$\pm$0.0106}} &
  0.4307 \tiny{$\pm$0.0691} &
  0.6794 \tiny{$\pm$0.1110} &
  {\color[HTML]{3531FF} \textbf{0.5629 \tiny{$\pm$0.0671}}} \\
XGBoost &
  \textbf{0.3503 \tiny{$\pm$0.0020}} &
  0.4325 \tiny{$\pm$0.0008} &
  0.7277 \tiny{$\pm$0.0004} &
  0.5405 \tiny{$\pm$0.0049} &
  0.4156 \tiny{$\pm$0.0087} &
  \textbf{0.7665 \tiny$\pm$0.0121} &
  0.5133 \tiny{$\pm$0.010} \\
CatBoost &
  0.3388 \tiny{$\pm$0.0006} &
  0.4312 \tiny{$\pm$0.0000} &
  \textbf{0.7328 \tiny{$\pm$0.0012}} &
  0.4806 \tiny{$\pm$0.0102} &
  0.2565 \tiny{$\pm$0.0494} &
  {\color[HTML]{3531FF} 0.7738 \tiny$\pm$0.0175} &
  0.5106 \tiny{$\pm$0.0377} \\ \hline
\multicolumn{1}{c}{\textbf{model}} &
  \textbf{dresses-sales} &
  \textbf{fars} &
  \textbf{jasmine} &
  \textbf{kick} &
  \textbf{okcupid-stem} &
  \textbf{online-shoppers} &
  \textbf{shrutime} \\ \hline
GFTab  &
  0.4252 \tiny$\pm$0.1100 &
  {\color[HTML]{FE0000} 0.6039 \tiny$\pm$0.0021} &
  0.7133 \tiny{$\pm$0.0194} &
  {\color[HTML]{FE0000} 0.5180 \tiny$\pm$0.0066} &
  {\color[HTML]{3531FF} 0.4122 \tiny$\pm$0.0210} &
  {\color[HTML]{29261B} \textbf{0.7366 \tiny{$\pm$0.0042}}} &
  \textbf{0.6641 \tiny{$\pm$0.0016}} \\
GRANDE &
  \textbf{0.4824 \tiny{$\pm$0.0656}} &
  \textbf{0.5575 \tiny{$\pm$0.0296}} &
  \textbf{0.7361 \tiny{$\pm$0.0212}} &
  {\color[HTML]{3531FF} 0.5175 \tiny$\pm$0.0006} &
  {\color[HTML]{FE0000} 0.4783 \tiny$\pm$0.0050} &
  0.6635 \tiny{$\pm$0.0112} &
  {\color[HTML]{FE0000} 0.6866 \tiny$\pm$0.0263} \\
TabPFN &
  0.3671 \tiny{$\pm$0.0002} &
  0.3828 \tiny{$\pm$0.0137} &
  0.7183 \tiny{$\pm$0.0031} &
  0.4749 \tiny{$\pm$0.0020} &
  0.3336 \tiny{$\pm$0.0052} &
  {\color[HTML]{3531FF} 0.7570 \tiny$\pm$0.0033} &
  0.5744 \tiny{$\pm$0.0101} \\
SCARF &
  0.4245 \tiny{$\pm$0.1140} &
  0.1366 \tiny{$\pm$0.0211} &
  0.5925 \tiny{$\pm$0.0394} &
  0.4751 \tiny{$\pm$0.0006} &
  0.2922 \tiny{$\pm$0.0117} &
  0.4631 \tiny{$\pm$0.0044} &
  0.4439 \tiny{$\pm$0.0014} \\
SubTab &
  {\color[HTML]{FE0000} 0.6184 \tiny$\pm$0.0105} &
  0.4203 \tiny{$\pm$0.0009} &
  0.5973 \tiny{$\pm$0.0069} &
  \textbf{0.4901 \tiny{$\pm$0.0013}} &
  \textbf{0.3728 \tiny{$\pm$0.0316}} &
  0.6404 \tiny{$\pm$0.0068} &
  0.6419 \tiny$\pm$0.0037 \\
VIME &
  0.4388 \tiny{$\pm$0.1270} &
  {\color[HTML]{3531FF} 0.5588 \tiny$\pm$0.0148} &
  {\color[HTML]{3531FF} 0.7394 \tiny$\pm$0.0108} &
  0.4524 \tiny{$\pm$0.0015} &
  0.3198 \tiny{$\pm$0.0292} &
  0.4601 \tiny{$\pm$0.0033} &
  0.5718 \tiny{$\pm$0.0397} \\
XGBoost &
  {\color[HTML]{3531FF} 0.5257 \tiny{$\pm$0.0169}} &
  0.4195 \tiny{$\pm$0.0003} &
  0.7357 \tiny{$\pm$0.0152} &
  0.4748 \tiny{$\pm$0.0000} &
  0.3063 \tiny{$\pm$0.0000} &
  {\color[HTML]{FE0000} 0.7791 \tiny$\pm$0.0050} &
  {\color[HTML]{3531FF} 0.6830 \tiny$\pm$0.0051} \\
CatBoost &
  0.3838 \tiny{$\pm$0.0290} &
  0.3571 \tiny{$\pm$0.001} &
  {\color[HTML]{FE0000} 0.7494 \tiny$\pm$0.0016} &
  0.4749 \tiny{$\pm$0.0000} &
  0.3036 \tiny{$\pm$0.0002} &
  0.7284 \tiny{$\pm$0.0020} &
  0.6334 \tiny{$\pm$0.0045} \\ \hline
\multicolumn{8}{l}{} \\ \hline
\multicolumn{8}{l}{\textbf{Panel B. Datasets with more continouse variables}} \\
\multicolumn{1}{c}{\textbf{model}} &
  \textbf{KDD} &
  \textbf{Shipping} &
  \textbf{churn} &
  \textbf{eye-movements} &
  \textbf{nomao} &
  \textbf{qsar} &
  \textbf{road-safety} \\ \hline
GFTab  &
  {\color[HTML]{3531FF} 0.7774 \tiny{$\pm$0.0164}} &
  \textbf{0.6203 \tiny{$\pm$0.0023}} &
  {\color[HTML]{3531FF} 0.6285 \tiny$\pm$0.0041} &
  \textbf{0.5600 \tiny{$\pm$0.0129}} &
  \textbf{0.8866 \tiny{$\pm$0.0006}} &
  0.7099 \tiny{$\pm$0.0064} &
  {\color[HTML]{FE0000} 0.7349 \tiny$\pm$0.0019} \\
GRANDE &
  0.7578 \tiny{$\pm$0.0126} &
  0.6036 \tiny{$\pm$0.0046} &
  0.5806 \tiny{$\pm$0.0091} &
  0.4737 \tiny{$\pm$0.0874} &
  0.8602 \tiny{$\pm$0.0065} &
  0.6714 \tiny{$\pm$0.0374} &
  \textbf{0.7330 \tiny{$\pm$0.0023}} \\
TabPFN &
  0.7090 \tiny{$\pm$0.0034} &
  {\color[HTML]{FE0000} 0.6411 \tiny$\pm$0.0028} &
  0.4624 \tiny{$\pm$0.0000} &
  0.5575 \tiny{$\pm$0.0058} &
  0.7818 \tiny{$\pm$0.0131} &
  \textbf{0.7716 \tiny{$\pm$0.0122}} &
  0.6960 \tiny{$\pm$0.0018} \\
SCARF &
  0.4969 \tiny{$\pm$0.0429} &
  0.5569 \tiny{$\pm$0.1136} &
  0.4624 \tiny{$\pm$0.0000} &
  0.4841 \tiny{$\pm$0.0166} &
  0.5293 \tiny{$\pm$0.0149} &
  0.6909 \tiny{$\pm$0.0366} &
  0.4993 \tiny{$\pm$0.0115} \\
SubTab &
  0.6135 \tiny{$\pm$0.0057} &
  0.5403 \tiny{$\pm$0.0011} &
  \textbf{0.5855 \tiny{$\pm$0.0156}} &
  0.5502 \tiny{$\pm$0.0029} &
  0.8019 \tiny{$\pm$0.0082} &
  0.6712 \tiny{$\pm$0.0218} &
  0.6624 \tiny{$\pm$0.0007} \\
VIME &
  0.6760 \tiny{$\pm$0.0066} &
  0.5941 \tiny{$\pm$0.0411} &
  {\color[HTML]{FE0000} 0.6422 \tiny$\pm$0.0266} &
  0.5260 \tiny{$\pm$0.0308} &
  {\color[HTML]{3531FF} 0.8929 \tiny$\pm$0.0032} &
  {\color[HTML]{FE0000} 0.7852 \tiny$\pm$0.0044} &
  0.7168 \tiny{$\pm$0.0012} \\
XGBoost &
  \textbf{0.7708 \tiny$\pm$0.0099} &
  0.5983 \tiny{$\pm$0.0028} &
  0.5569 \tiny{$\pm$0.0252} &
  {\color[HTML]{3531FF} 0.5641 \tiny$\pm$0.0112} &
  {\color[HTML]{FE0000} 0.8996 \tiny$\pm$0.0012} &
  0.7318 \tiny{$\pm$0.0259} &
  {\color[HTML]{3531FF} 0.7334 \tiny$\pm$0.0128} \\
CatBoost &
  {\color[HTML]{FE0000} 0.7982 \tiny$\pm$0.0068} &
  {\color[HTML]{3531FF} 0.6317 \tiny$\pm$0.0045} &
  0.4720 \tiny{$\pm$0.0166} &
  {\color[HTML]{FE0000} 0.5921 \tiny$\pm$0.0169} &
  0.8803 \tiny$\pm$0.0006 &
  {\color[HTML]{3531FF} 0.7830 \tiny$\pm$0.0110} &
  0.7233 \tiny{$\pm$0.0068} \\ \hline
\end{tabular}}
    \caption{Comparison of F1 score between GFTab and baseline models on 21 tabular benchmark datasets in 20\% labeled training with 20\% label noise. The best performing method is highlighted in \textcolor{red}{red} and the second best in \textcolor{blue}{blue}, while the third best is \textbf{bold}.}
    \label{f1_12}
    \vspace{-0.5cm}
\end{table*}

{\bfseries Baseline models \& hyperparameter.} We use GRANDE \cite{marton2024grande}, TabPFN \cite{hollmann2023tabpfn} SCARF \cite{bahri2021scarf}, VIME \cite{yoon2020vime}, SubTab \cite{ucar2021subtab}, XGBoost \cite{chen2016xgboost} and CatBoost \cite{prokhorenkova2018catboost} as baseline models. For XGBoost and CatBoost, we performed hyperparameter optimization using Optuna \cite{akiba2019optuna}, conducting 250 experiments for each model. The search space for hyperparameters was kept consistent with the one used in GRANDE \cite{marton2024grande} to ensure a fair comparison. Regarding the deep learning models, such as GRANDE, TabPFN, SCARF, VIME, and SubTab, we provide detailed information about the hyperparameter tuning process in Appendix B. 

{\bfseries Evaluation metrics.}  For efficient presentation in our ablation study, we define a \textit{win matrix}. A win matrix is a $k$ by $k$ matrix, where $k$ is the number of method being compared in this study. The entry $W_{i,j}$ is calculated as the proportion of the number of times method $i$ outperformed method $j$ over $m$ datasets. It is represented as follows:
\vspace{-0.2cm}
\begin{equation}
    \small
    W_{i,j} = \sum_{d=1}^{m} \mathbb{I}[\text{method}\, i\, \text{beats method}\, j \,\text{on dataset}\, d] / m
\end{equation}

\subsection{Is GFTab really effective for tabular datasets?}
We evaluate GFTab's performance on 21 diverse benchmark datasets, comparing it against seven baseline models under two primary conditions: 20\% labeled training data (Table \ref{f1_02}) and 20\% labeled training data with 20\% label noise (Table \ref{f1_12}). The datasets are divided into two panels: Panel A for datasets with more categorical variables and Panel B for those with more continuous variables.

In Table \ref{f1_02}, which presents results for the 20\% labeled data settings, GFTab demonstrates superior performance across diverse datasets. It achieves the highest F1 score in nearly half of the cases (10 out of 21) and ranks among the top three in an additional seven. GFTab particularly excels with categorical-dominant datasets (Panel A), leading in 7 out of 14 cases and placing in the top three for 5 more. For continuous-dominant datasets (Panel B), GFTab maintains strong performance, topping 3 out of 7 cases and ranking in the top three for 2 others. Table \ref{f1_12} shows GFTab's robustness when 20\% label noise is introduced. It retains competitive performance overall, securing the highest F1 score in 8 datasets and ranking among the top three in 10 others. Notably, GFTab's resilience is particularly evident in datasets with a higher proportion of categorical variables, even under these noisy conditions.

While certain methods perform consistently well on some datasets (SubTab on dresses-sales and qsar, CatBoost on jasmine and KDD, GRANDE on okcupid-stem and shrutime, XGboost on online-shoppers), we emphasize that there is no one-size-fits-all solution for all tabular datasets \cite{marton2024grande}. Our proposed GFTab, while not universally superior on all datasets, clearly performs robustly regardless of label noise when compared to the top three performers. Also, we found that a label percentage of 10\% yielded similar results to the 20\% setting. Detailed results for this additional experiment can be found in Appendix E.

\subsection{How to corrupt categorical variables effectively?} \label{exp_corruption_methods}
When designing corruption methods for categorical variables, we must consider their ordered or unordered nature. For ordered variables, while precise quantification of distances between values is impossible, we can still determine relative closeness, allowing corruption to neighboring values (GFTab). Unordered variables lack this concept of similarity, requiring random replacement with another category (Permute). However, challenges arise with imbalanced distributions, as categorical variables can exhibit identical discrete values across different items, complicating the creation of \textit{views}.To address this, we can employ a row-wise random selection strategy (Random). Unfortunately, it is challenging to choose one of the aforementioned methods separately for each categorical column based on its characteristics. Therefore, an alternative solution is to project the categorical variable into the embedding space and corrupt it in this space, regardless of the type of the categorical variable (Embed).

\begin{figure}[h!] 
  \centering
  \includegraphics[width=\linewidth]{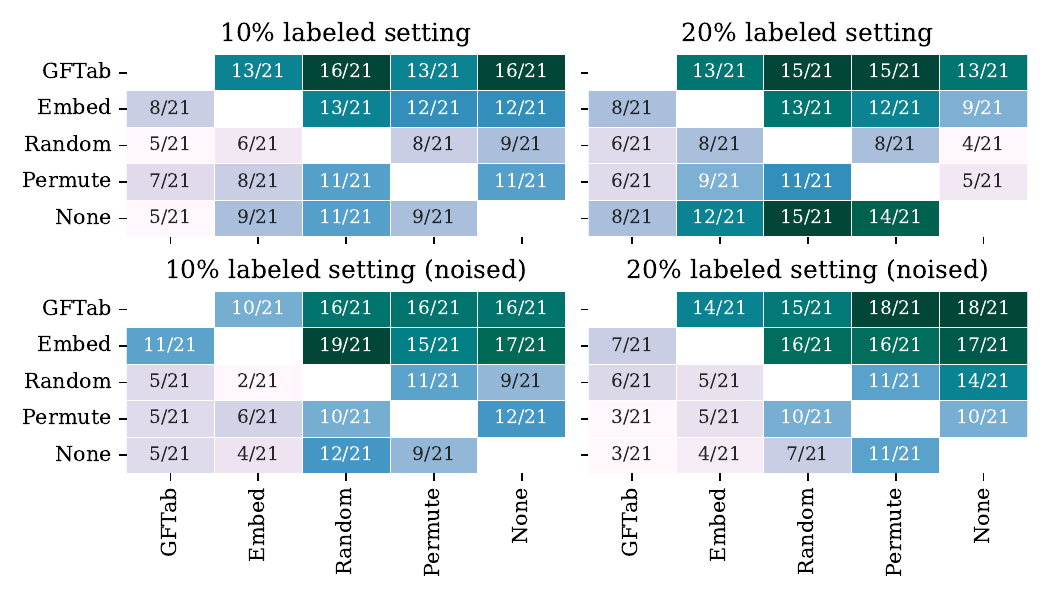}
  \caption{Win matrices for different categorical variable corruption methods.}
  \label{figure_cat}
\vspace{-20pt}
\end{figure}

To evaluate the effectiveness of these approaches, we tested five methods on our benchmark datasets: GFTab, Embed, Random, Permute, and None (where None indicates no categorical feature corruptions). The results, as shown in Figure \ref{figure_cat}, demonstrate the efficacy of our corruption methods for categorical variables, consistently outperforming other methods. Notably, GFTab shows marked performance enhancement compared to Permute and Random, particularly in settings with noisy labels. This provides compelling evidence that our selected corruption method is beneficial in handling categorical variables within tabular datasets.

\subsection{Is geodesic flow useful for tabular datasets?}
\begin{figure}[htbp!] 
  \centering
  \includegraphics[width=\linewidth]{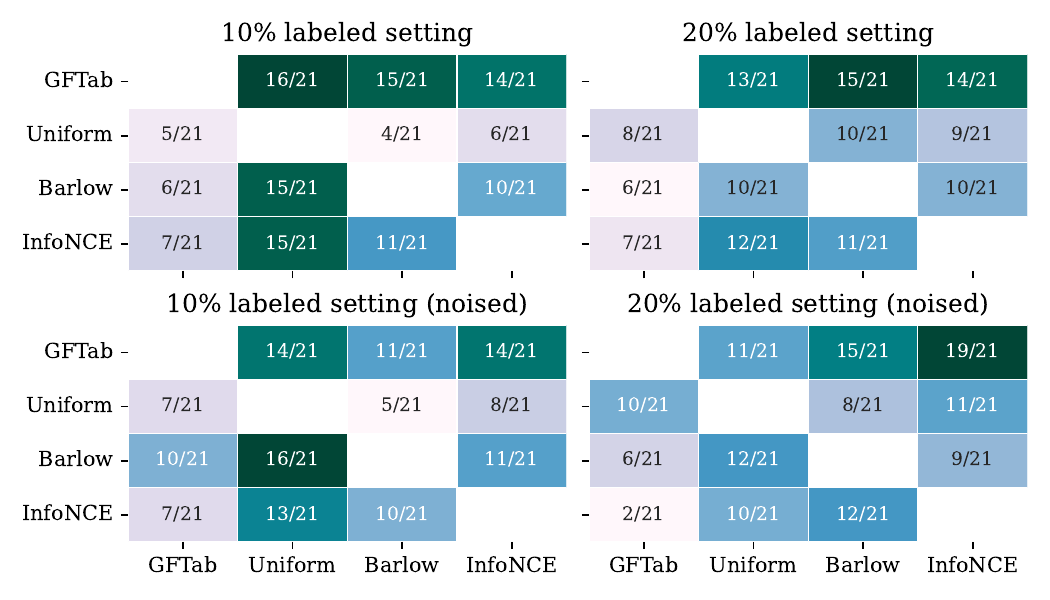}
  \caption{Win matrix between GFTab with three different similarity losses.}
  \label{figure_loss}
  \vspace{-0.50cm}
\end{figure}

In this section, we explore the effect of different loss functions on the performance of GFTab. The \textit{win-matrix} in Figure \ref{figure_loss} shows that GFTab loss function had the most number of wins compared to the other loss functions (Uniform-Alignment \cite{wang20k} Barlow Twins \cite{barlow},  and InfoNCE \cite{oord2018representation}). In the absence of label noise, it surpassed the other functions on at least 14 out of 21 datasets. Notably, GFTab outperformed these alternative loss functions regardless of label noise.

The superiority of GFTab loss function over traditional methods may be attributed to its ability to capture the geometric relationships between data representations in a more nuanced and comprehensive manner. Traditional methods often rely on Euclidean space metrics, which may not adequately capture the complex relationships present in high-dimensional tabular data. In contrast, the geodesic flow kernel provides a continuous and smooth transition between subspaces, allowing GFTab to capture subtle variations in the data that might be missed by traditional approaches. This nature may enables GFTab to more accurately model and learn from the intrinsic structure of tabular datasets.
\subsection{How to choose the balance between $\mathcal{L}_{\text{ce}}$ and $\mathcal{L}_{\text{sim}}$?}
\begin{figure}[h!] 
  \centering
  \includegraphics[width=\linewidth]{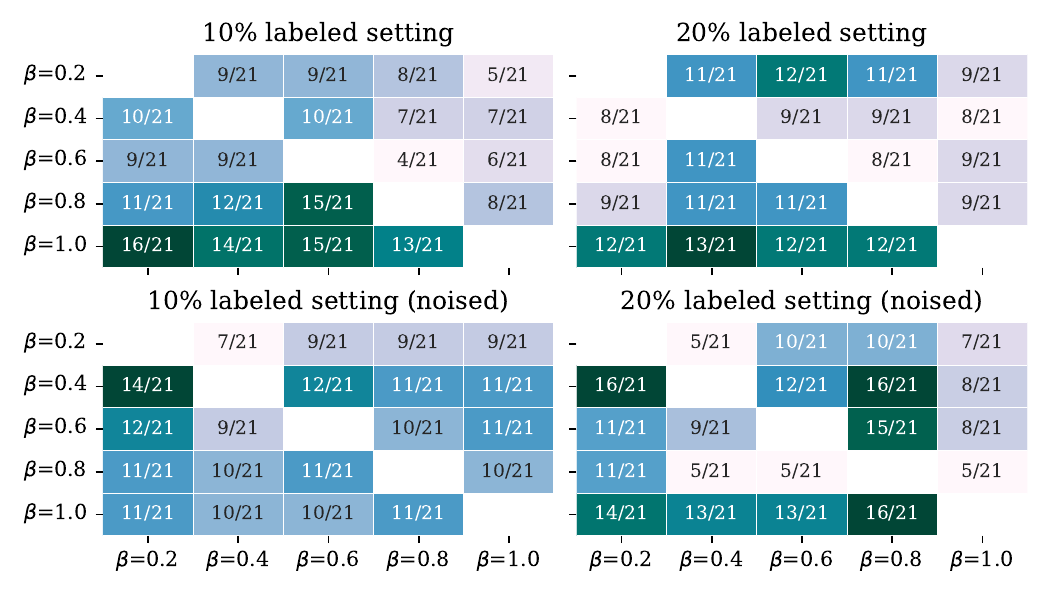}
  \caption{Performance of GFTab with different levels of balance parameter}
  \label{figure_balance_parameter}
\vspace{-24pt}
\end{figure}

The GFTab is based on semi-supervised learning so that it is trained to minimize  $\mathcal{L}_{\text{GFTab}}= \mathcal{L}_{\text{sim}}$+$\beta \mathcal{L}_{\text{ce}}$. we compared the performance of the model across various ranges of $\beta$. As a result, $\beta$ = 1.0 yielded the best balance. 
As shown in Figure \ref{figure_balance_parameter}, we conducted an analysis of the impact of different $\beta$ values on GFTab's performance across various labeling scenarios. The results show that $\beta$ = 1.0 yields good performance across different settings.

In the 10\% labeled setting, $\beta$ = 1.0 outperformed other $\beta$ values on a majority of datasets. This trend continued for the 20\% labeling setting, although slightly weaker, with $\beta$ = 1.0 maintaining competitive performance. Even in the presence of label noise, $\beta$ = 1.0 showed robustness, particularly in the 20\% labeled setting with noise. However, the 10\% labeled setting did not show a significant difference between $\beta$, but performed a little better at $\beta$ = 0.4. 

While the optimal $\beta$ might vary slightly depending on specific dataset characteristics or labeling conditions, our experiments indicate that $\beta$ = 1.0 is a reliable choice that performs well across a range of settings.

\section{Conclusion}
GFTab is a semi-supervised framework for mixed-variable tabular datasets. It incorporates three key components: variable-specific corruption methods, similarity loss based on geodesic flow kernel, and tree-based embedding. This approach enables effective learning from both labeled and unlabeled data, addressing challenges specific to tabular data structures. Experiments across 21 diverse benchmark datasets show GFTab consistently outperforming existing ML/DL models, particularly in scenarios with limited labeled data or label noise. GFTab's ability to handle tabular data characteristics makes it a robust and versatile solution applicable to various domains.

\section{Acknowledgments}
This work was supported by National Research Foundation of Korea (NRF) grant (No. NRF-2022R1I1A4069163 and No. RS-2024-00340339) and Institute of Information \& Communications Technology Planning \& Evaluation(IITP) grant (No. RS-2020-II201336, Artificial Intelligence Graduate School Program(UNIST)) funded by the Korea government(MSIT)

\bibliography{aaai25}
\clearpage

\clearpage
\label{ch:appendix}
\newpage
\newpage

\appendix
\begin{center}
\huge\bfseries
Supplementary Materials    
\end{center}

\subsection{A. Benchmark datasets} \label{appendix:A1}
Table \ref{APPENDIX:DATA} lists 21 datasets configured following the criteria explained in Section Tabular Benchmarks Dataset. The table shows each dataset's type, name, sample size, number of categorical and continuous features and number of classes.


\begin{table}[htb!]
\begin{center}
\tiny 

\begin{tabular}{clcccc} 
\hline
Type                & Dataset name      & \# Samples & \# Cat & \# Cont & \# class \\ \hline
\multirow{10}{*}{$\text{M}_{\text{cat}} \ge \text{M}_{\text{cont}}$} & cmc               & 1,325      & 7      & 2       & 3        \\
                    & jasmine           & 2,685      & 136    & 8       & 2        \\
                    & credit-g          & 900        & 13     & 7       & 2        \\
                    & Online\_shoppsers & 9,864      & 13     & 4       & 2        \\
                    & dresses-sales     & 450        & 11     & 1       & 2        \\
                    & shrutime          & 8,000      & 6      & 4       & 2        \\
                    & credit-approval   & 590        & 9      & 6       & 2        \\
                    & Insurance         & 18,838     & 7      & 3       & 2        \\
                    & bank              & 40,689     & 9      & 7       & 2        \\
                    & fars              & 90,863     & 15     & 14      & 7        \\
                    & okcupid\_stem     & 23,966     & 11     & 2       & 3        \\
                    & adult             & 43,957     & 8      & 6       & 2        \\
                    & Diabetes          & 91,589     & 36     & 11      & 3        \\
                    & kick              & 60,499     & 16     & 14      & 2        \\ \hline
\multirow{7}{*}{$\text{M}_{\text{cat}} \le \text{M}_{\text{cont}}$} & eye\_movement     & 6,848      & 3      & 20      & 2        \\
                    & Shipping          & 8,799      & 4      & 5       & 2        \\
                    & KDD               & 4,528      & 11     & 34      & 2        \\
                    & churn             & 4,500      & 4      & 16      & 2        \\
                    & qsar              & 844        & 10     & 30      & 2        \\
                    & nomao             & 31,018     & 29     & 90      & 2        \\
                    & road-safety       & 100,585    & 3      & 29      & 2       \\ \hline
\end{tabular}
\caption{Dataset description.}
\label{APPENDIX:DATA}
\end{center}
\end{table}
%


\subsection{B. Hyperparameters tune} \label{appendix:B1} Table \ref{hyper} shows the hyperparameter search ranges for GKSMT and ML/DL, which were optimized using Optuna \cite{akiba2019optuna} or grid-search, respectively. And, we conducted on a one NVIDIA RTX 3090 24GB GPUs. Also, all experiments are trained using the AdamW optimizer \cite{loshchilov2017decoupled} with an initial learning rate of $2 \times 10^{-3}$ and early stopping within 10 epochs and a learning rate schedule that uses a stepLR decay with optional warmup.

\begin{table}[h!]
\centering
    \centering
{\fontsize{6.5}{7.0}\selectfont
\begin{tabular}{cc}
\hline
\multicolumn{2}{c}{\textbf{Panel A. GKSMT}}           \\
Parameter                   & Distribution            \\ \hline
\text{d}\_{\text{emb}}      & [12,24,36,48]           \\
dim\_head                   & [12,24,36,48]           \\
layers                      & [1,2,4]                 \\
depths                      & [1,2,4]                 \\ \hline
\multicolumn{2}{l}{}                                  \\ \hline
\multicolumn{2}{c}{\textbf{Panel B. GRANDE}}          \\
Parameter                   & Distribution            \\ \hline
depth                       & [4,6]                   \\
n\_estimators               & [512,1024]              \\
dropout                     & [0.00, 0.25, 0.5]       \\
learning\_rate\_weights     & LogUniform[0.0001, 0.1] \\
learning\_rate\_index       & LogUniform[0.005, 0.2]  \\
learning\_rate\_values      & LogUniform[0.005, 0.2]  \\
learning\_rate\_leaf        & LogUniform[0.005, 0.2]  \\ \hline
\multicolumn{2}{l}{}                                  \\ \hline
\multicolumn{2}{c}{\textbf{Panel C. TabPFN}}          \\
Parameter                   & Distribution            \\ \hline
layers                      & [1, 2]                  \\
activation\_function         & [Tanh, Leaky ReLU, ELU] \\
N\_ensemble\_configurations & [4, 8, 12, 24]          \\ \hline
\multicolumn{2}{l}{}                                  \\ \hline
\multicolumn{2}{c}{\textbf{Panel D. SCARF}}           \\
Parameter                   & Distribution            \\ \hline
hidden\_dim                 & [64,128,256,512]        \\
Layers                      & [1,3,5]                 \\
head\_dim                   & [64,128,256]            \\
corruption\_rate            & [0.1,0.3,0.5,0.7]       \\ \hline
\multicolumn{2}{l}{}                                  \\ \hline
\multicolumn{2}{c}{\textbf{Panel E. SubTab}}          \\
Parameter                   & Distribution            \\ \hline
mask\_ratio                 & [0.15, 0.2]             \\
encoder\_layers             & [1, 2, 3]               \\
dncoder\_layers             & [1, 2, 3]               \\
encoder\_dim                & [512, 784, 1024]        \\
decoder\_dim                & [512, 784, 1024]        \\ \hline
\multicolumn{2}{l}{}                                  \\ \hline
\multicolumn{2}{c}{\textbf{Panel F. VIME}}            \\
Parameter                   & Distribution            \\ \hline
hidden\_dim                 & [64,128,256,512]        \\
alpha                       & [0.5,1.0,1.5,2.0]       \\
$\mathrm{K}$                & [1,3,5,7]               \\
beta                        & [0.1,0.3,0.5]           \\ \hline
\multicolumn{1}{l}{}        & \multicolumn{1}{l}{}    \\ \hline
\multicolumn{2}{c}{\textbf{Panel G. XGBoost}}         \\
Parameter                   & Distribution            \\ \hline
max\_depths                 & randInt[2,12]           \\
alpha                       & logUniform[1e-8,1.0]    \\
lambda                      & logUniform[1e-8,1.0]    \\
eta                         & logUniform[0.01, 0.3]   \\
learning\_rate              & logUniform[0.01, 0.3]   \\
n\_estimators                & randInt[500,1000]       \\ \hline
\multicolumn{1}{l}{}        & \multicolumn{1}{l}{}    \\ \hline
\multicolumn{2}{c}{\textbf{Panel G. CatBoost}}        \\
Parameter                   & Distribution            \\ \hline
max\_depths                 & randInt[2,12]           \\
l2\_leaf\_reg               & logUniform[0.5, 30]     \\
learning\_rate              & logUniform[0.01, 0.3]   \\
n\_estimators                & randInt[500,1000]       \\ \hline
\end{tabular}}

    \caption{GKSMT and ML/DL models hyperparameter search ranges.}
    \label{hyper}
\end{table}

\subsection{C. The effect of VSN} \label{appendix_8}
In Table \ref{APPENDIX:C}, Variable Selection with Noise (VSN) demonstrates meaningful performance improvements across diverse tabular datasets. With an average gain of 2.76$\pm$3.45\%, our experiments reveal particularly strong results for healthcare data, with the Diabetes dataset showing an 8.77\% improvement.
Eight out of nine datasets exhibited positive gains, with four datasets showing improvements above 3\%. The method delivers consistent enhancements across varied applications, from road safety (4.95\%) to complex competition datasets like kick (4.20\%). While the qsar dataset showed a 4.02\% decrease, this is outweighed by the consistent positive improvements across other domains. The method's robust performance across different data types and sizes suggests its effectiveness in real-world scenarios where data quality varies.

\begin{table}[h!]
\centering
    \centering
{\fontsize{6.5}{7.0}\selectfont
\begin{tabular}{lccc}
\hline
        \textbf{Dataset} & \textbf{GFTab(W noise)} & \textbf{Del\_impVar} & \textbf{Improvement(\%)} \\ \hline
        Diabetes       & 0.3740 & 0.3412 & 8.770 \\
        fars           & 0.5693 & 0.5577 & 2.037 \\
        kick           & 0.4879 & 0.4674 & 4.201 \\
        KDD            & 0.7717 & 0.7608 & 1.412 \\
        churn          & 0.6153 & 0.6099 & 0.877 \\
        eye-movements  & 0.5321 & 0.5145 & 3.307 \\
        nomao          & 0.8312 & 0.8034 & 3.344 \\
        qsar           & 0.6472 & 0.6732 & -4.017 \\
        road-safety    & 0.7377 & 0.7012 & 4.947 \\ \hline
\label{APPENDIX:D}
\end{tabular}}

    \caption{Comparison of GFTab vs important variables excluded and their improvement}
    \label{APPENDIX:C}
\end{table}

\subsection{D. The effect of tree-based embedding} \label{appendix_9}
Tree-based embedding demonstrates compelling performance improvements in tabular data processing. In Table \ref{APPENDIX:D}, Our experiments with 20\% label noise reveal an average accuracy gain of 4.31$\pm$3.56\% across diverse datasets. Notably, the approach achieved remarkable improvements of 8.83\% and 8.24\% on the qsar and credit-approval datasets, respectively.
The results are particularly impressive for complex financial and chemical datasets, where accurate predictions are crucial. Seven out of nine tested datasets showed positive improvements, with six displaying gains exceeding 4\%. This consistent performance enhancement across varied domains suggests that tree-based embedding effectively captures important structural relationships in tabular data.
While two datasets (dresses-sales and shrutime) showed marginal performance decreases (-0.64\% and -0.24\%), these slight variations are outweighed by the substantial gains observed across the majority of cases.

\begin{table}[h!]
\centering
    \centering
{\fontsize{6.5}{7.0}\selectfont
\begin{tabular}{lccc}
\hline
        \textbf{Dataset} & \textbf{GFTab(W tree)} & \textbf{GFTab(W/o Tree)} & \textbf{Improvement(\%)} \\ \hline
        cmc              & 0.4598  & 0.4290  & 6.699 \\
        credit-approval  & 0.7853  & 0.7206  & 8.239 \\
        dresses-sales    & 0.4252  & 0.4279  & -0.635 \\
        jasmine          & 0.7133  & 0.6750  & 5.369 \\
        online-shoppers  & 0.7366  & 0.7012  & 4.806 \\
        shrutime         & 0.6641  & 0.6657  & -0.241 \\
        KDD              & 0.7774  & 0.7717  & 0.733 \\
        eye-movements    & 0.5600  & 0.5321  & 4.982 \\
        qsar             & 0.7099  & 0.6472  & 8.832 \\ \hline
\label{APPENDIX:D}
\end{tabular}}

    \caption{Comparison of GFTab (with tree) vs GFTab (without tree) and their improvement.}
    \label{APPENDIX:D}
\end{table}

\subsection{E. Is GFTab really effective for tabular datasets? (10\% labeled settings)} \label{appendix_10}

Lastly, we present the performance of GFTab and the baseline models in terms of 10\% labeled setting. Tables \ref{f1_01} and \ref{f1_11} summarize the results for the 10\% labeled training data and 10\% labeled training data with 10\% label noise, respectively.

The results for the 10\% labeled setting further reinforce GFTab's effectiveness across diverse tabular datasets. In Table \ref{f1_01}, which presents results for the 10\% labeled data scenario, GFTab demonstrates strong performance, achieving the highest F1 score in 9 out of 21 datasets and ranking among the top three in an additional 6 datasets. GFTab's performance is particularly notable in datasets with a higher proportion of categorical variables (Panel A), where it leads in 6 out of 14 cases and places in the top three for 4 more. For datasets dominated by continuous variables (Panel B), GFTab maintains competitive performance, topping 3 out of 7 cases and ranking in the top three for 2 others.

Table \ref{f1_11} showcases GFTab's robustness when 10\% label noise is introduced to the 10\% labeled data. Even under these challenging conditions, GFTab retains competitive performance overall, securing the highest F1 score in 5 datasets and ranking among the top three in 8 others. Notably, GFTab's resilience is particularly evident in datasets with a higher proportion of categorical variables, maintaining strong performance even under noisy conditions.

These results align closely with the findings from the 20\% labeled setting, indicating that GFTab's effectiveness is consistent across different proportions of labeled data. The model's ability to perform well with even less labeled data (10\% and 20\%) further highlights its efficiency in leveraging both labeled and unlabeled data in semi-supervised learning scenarios. This consistency in performance across different labeling ratios underscores GFTab's robustness and versatility in handling tabular datasets, particularly in real-world scenarios where labeled data may be scarce or noisy.

\begin{table*}[!htbp]
\centering
    \centering
{\fontsize{6.3}{6.8}\selectfont
\begin{tabular}{lccccccc}
\hline
\multicolumn{8}{l}{\textbf{Panel A. Datasets with more categorical variables}} \\
\multicolumn{1}{c}{\textbf{model}} &
  \textbf{Diabetes} &
  \textbf{Insurance} &
  \textbf{adult} &
  \textbf{bank} &
  \textbf{cmc} &
  \textbf{credit-approval} &
  \textbf{credit-g} \\ \hline
GFTab &
  {\color[HTML]{FF0000} 0.3722 \tiny{$\pm$0.0059}} &
  {\color[HTML]{FF0000} 0.4418 \tiny$\pm$0.0116} &
  {\color[HTML]{FF0000} 0.7914 \tiny$\pm$0.0013} &
  {\color[HTML]{FF0000} 0.7224 \tiny$\pm$0.010} &
  0.4403 \tiny{$\pm$0.0011} &
  {\color[HTML]{3531FF} 0.7908 \tiny$\pm$0.0212} &
  \textbf{0.5350 \tiny$\pm$0.0783} \\
GRANDE &
  {\color[HTML]{3531FF} 0.3515 \tiny$\pm$0.0032} &
  0.4225 \tiny{$\pm$0.0001} &
  \textbf{0.7559 \tiny{$\pm$0.0028}} &
  \textbf{0.7066 \tiny$\pm$0.0104} &
  {\color[HTML]{3531FF} 0.4517 \tiny$\pm$0.0192} &
  {\color[HTML]{FF0000} 0.8109 \tiny$\pm$0.0097} &
  {\color[HTML]{FF0000} 0.5968 \tiny$\pm$0.0236} \\
TabPFN &
  0.2551 \tiny{$\pm$0.0028} &
  0.4312 \tiny{$\pm$0.0001} &
  {\color[HTML]{3531FF} 0.7686 \tiny$\pm$0.0039} &
  0.6624 \tiny{$\pm$0.0012} &
  \textbf{0.4501 \tiny$\pm$0.0027} &
  \textbf{0.7568 \tiny$\pm$0.0118} &
  0.4118 \tiny{$\pm$0.0000} \\
SCARF &
  0.2335 \tiny{$\pm$0.0000} &
  0.4312 \tiny{$\pm$0.0001} &
  0.4368 \tiny{$\pm$0.0187} &
  0.4688 \tiny{$\pm$0.0001} &
  0.3192 \tiny{$\pm$0.0209} &
  0.6036 \tiny{$\pm$0.0922} &
  0.4351 \tiny{$\pm$0.0298} \\
SubTab &
  0.3012 \tiny{$\pm$0.0011} &
  \textbf{0.4313 \tiny{$\pm$0.0004}} &
  0.7285 \tiny{$\pm$0.0067} &
  0.6139 \tiny{$\pm$0.0091} &
  0.4425 \tiny{$\pm$0.0391} &
  0.6734 \tiny{$\pm$0.0786} &
  0.5321 \tiny{$\pm$0.0212} \\
VIME &
  0.3480 \tiny$\pm$0.0313 &
  0.4311 \tiny{$\pm$0.0001} &
  0.7664 \tiny$\pm$0.0057 &
  0.6496 \tiny{$\pm$0.0179} &
  {\color[HTML]{FF0000} 0.5228 \tiny$\pm$0.0336} &
  0.7468 \tiny{$\pm$0.1637} &
  {\color[HTML]{3531FF} 0.5723 \tiny$\pm$0.0891} \\
XGBoost &
  \textbf{0.3494 \tiny$\pm$0.0001} &
  {\color[HTML]{3531FF} 0.4325 \tiny$\pm$0.0017} &
  0.7281 \tiny{$\pm$0.0003} &
  {\color[HTML]{3531FF} 0.7110 \tiny$\pm$0.0100} &
  0.4478 \tiny{$\pm$0.0563} &
  0.7425 \tiny{$\pm$0.0284} &
  0.5089 \tiny{$\pm$0.0629} \\
CatBoost &
  0.3474 \tiny{$\pm$0.0005} &
  0.4312 \tiny{$\pm$0.0001} &
  0.6122 \tiny{$\pm$0.0001} &
  0.7011 \tiny{$\pm$0.0012} &
  0.3979 \tiny{$\pm$0.0372} &
  0.7486 \tiny{$\pm$0.0231} &
  0.4118 \tiny{$\pm$0.0001} \\ \hline
\multicolumn{1}{c}{\textbf{model}} &
  \textbf{dresses-sales} &
  \textbf{fars} &
  \textbf{jasmine} &
  \textbf{kick} &
  \textbf{okcupid-stem} &
  \textbf{online-shoppers} &
  \textbf{shrutime} \\ \hline
GFTab &
  0.4333 \tiny{$\pm$0.1147} &
  {\color[HTML]{FF0000} 0.5964 \tiny$\pm$0.0284} &
  \textbf{0.7868  \tiny{$\pm$0.0145}} &
  {\color[HTML]{FF0000} 0.5137 \tiny$\pm$0.0043} &
  0.3615 \tiny{$\pm$0.0137} &
  0.7422 \tiny$\pm$0.0138 &
  {\color[HTML]{3531FF} 0.7086 \tiny$\pm$0.0067} \\
GRANDE &
  \textbf{0.5265 \tiny{$\pm$0.0655}} &
  0.5053 \tiny{$\pm$0.0083} &
  0.7606 \tiny$\pm$0.0114 &
  0.4654 \tiny{$\pm$0.0001} &
  {\color[HTML]{3531FF} 0.4566 \tiny$\pm$0.0062} &
  0.7097 \tiny{$\pm$0.0034} &
  {\color[HTML]{FF0000} 0.7210 \tiny$\pm$0.0115} \\
TabPFN &
  0.3671 \tiny{$\pm$0.0000} &
  \textbf{0.5246 \tiny$\pm$0.0012} &
  0.6853 \tiny{$\pm$0.0001} &
  0.4749 \tiny{$\pm$0.0001} &
  0.3706 \tiny{$\pm$0.0009} &
  {\color[HTML]{3531FF} 0.7696 \tiny$\pm$0.0010} &
  \textbf{0.6871 \tiny$\pm$0.0049} \\
SCARF &
  {\color[HTML]{3531FF} 0.5401 \tiny$\pm$0.0404} &
  0.1297 \tiny{$\pm$0.0168} &
  0.5759 \tiny{$\pm$0.0375} &
  0.4747 \tiny{$\pm$0.0002} &
  0.2828 \tiny{$\pm$0.0074} &
  0.4588 \tiny{$\pm$0.0014} &
  0.4448 \tiny{$\pm$0.0025} \\
SubTab &
  {\color[HTML]{FF0000} 0.5892 \tiny$\pm$0.0218} &
  0.4215 \tiny{$\pm$0.0045} &
  0.5984 \tiny{$\pm$0.0175} &
  {\color[HTML]{3531FF} 0.4777 \tiny$\pm$0.0022} &
  0.3919 \tiny{$\pm$0.0069} &
  0.6204 \tiny{$\pm$0.0283} &
  0.6494 \tiny{$\pm$0.0122} \\
VIME &
  0.4600 \tiny{$\pm$0.0346} &
  {\color[HTML]{3531FF} 0.5721 \tiny$\pm$0.0041} &
  0.7365 \tiny{$\pm$0.0050} &
  \textbf{0.4771 \tiny{$\pm$0.0006}} &
  0.3178 \tiny{$\pm$0.0180} &
  0.4597 \tiny{$\pm$0.0015} &
  0.6800 \tiny{$\pm$0.0421} \\
XGBoost &
  0.4835 \tiny{$\pm$0.0057} &
  0.4192 \tiny{$\pm$0.0000} &
  {\color[HTML]{3531FF} 0.7884 \tiny$\pm$0.0146} &
  0.4758 \tiny{$\pm$0.0017} &
  {\color[HTML]{29261B} \textbf{0.4522 \tiny$\pm$0.0022}} &
  {\color[HTML]{FF0000} 0.7857 \tiny{$\pm$0.0091}} &
  0.6827 \tiny{$\pm$0.0119} \\
CatBoost &
  0.3938 \tiny{$\pm$0.0231} &
  0.3571 \tiny{$\pm$0.0000} &
  {\color[HTML]{FF0000} 0.7952 \tiny$\pm$0.0053} &
  0.4749 \tiny{$\pm$0.0000} &
  {\color[HTML]{FF0000} 0.4712 \tiny$\pm$0.0032} &
  \textbf{0.7608 \tiny$\pm$0.0077} &
  0.4775 \tiny{$\pm$0.0370} \\ \hline
\multicolumn{8}{l}{} \\ \hline
\multicolumn{8}{l}{\textbf{Panel B. Datasets with more continouse variables}} \\
\multicolumn{1}{c}{\textbf{model}} &
  \textbf{KDD} &
  \textbf{Shipping} &
  \textbf{churn} &
  \textbf{eye-movements} &
  \textbf{nomao} &
  \textbf{qsar} &
  \textbf{road-safety} \\ \hline
GFTab &
  {\color[HTML]{3531FF} 0.7988 \tiny{$\pm$0.0096}} &
  {\color[HTML]{FF0000} 0.7086 \tiny$\pm$0.0091} &
  {\color[HTML]{FF0000} 0.7500 \tiny$\pm$0.0397} &
  0.5420 \tiny{$\pm$0.0211} &
  {\color[HTML]{FF0000} 0.9279 \tiny$\pm$0.0057} &
  \textbf{0.7727 \tiny{$\pm$0.0040}} &
  0.7446 \tiny{$\pm$0.0067} \\
GRANDE &
  \textbf{0.7901 \tiny$\pm$0.0116} &
  0.6244 \tiny{$\pm$0.0053} &
  0.7043 \tiny{$\pm$0.0421} &
  0.5554 \tiny{$\pm$0.0080} &
  0.9020 \tiny{$\pm$0.0001} &
  0.7718 \tiny$\pm$0.0152 &
  {\color[HTML]{FF0000} 0.7561 \tiny$\pm$0.0021} \\
TabPFN &
  0.7565 \tiny{$\pm$0.0023} &
  {\color[HTML]{3531FF} 0.6513 \tiny$\pm$0.0017} &
  {\color[HTML]{3531FF} 0.7495 \tiny$\pm$0.0024} &
  {\color[HTML]{FF0000} 0.5803 \tiny$\pm$0.0047} &
  0.8803 \tiny{$\pm$0.0028} &
  {\color[HTML]{FF0000} 0.8336 \tiny$\pm$0.0000} &
  0.7335 \tiny{$\pm$0.0007} \\
SCARF &
  0.5238 \tiny{$\pm$0.0198} &
  0.6020 \tiny{$\pm$0.0177} &
  0.4670 \tiny{$\pm$0.005} &
  0.4877 \tiny{$\pm$0.0018} &
  0.5296 \tiny{$\pm$0.0235} &
  0.6571 \tiny{$\pm$0.0204} &
  0.5009 \tiny{$\pm$0.0031} \\
SubTab &
  0.5486 \tiny{$\pm$0.0651} &
  0.5347 \tiny{$\pm$0.0106} &
  0.6574 \tiny{$\pm$0.0548} &
  \textbf{0.5626 \tiny{$\pm$0.0069}} &
  \textbf{0.9015 \tiny{$\pm$0.0088}} &
  0.7593 \tiny{$\pm$0.0154} &
  0.6720 \tiny{$\pm$0.0010} \\
VIME &
  0.6907 \tiny{$\pm$0.0164} &
  0.6018 \tiny{$\pm$0.0148} &
  {\color[HTML]{29261B} \textbf{0.7259 \tiny$\pm$0.0304}} &
  0.5482 \tiny{$\pm$0.0277} &
  {\color[HTML]{3531FF} 0.9165 \tiny$\pm$0.0028} &
  {\color[HTML]{3531FF} 0.7901 \tiny$\pm$0.0158} &
  0.7120 \tiny{$\pm$0.0025} \\
XGBoost &
  {\color[HTML]{FF0000} 0.8057 \tiny$\pm$0.0002} &
  0.6384 \tiny{$\pm$0.0021} &
  0.5752 \tiny{$\pm$0.0160} &
  0.5295 \tiny{$\pm$0.0044} &
  0.9068 \tiny$\pm$0.0011 &
  0.7703 \tiny{$\pm$0.0294} &
  {\color[HTML]{3531FF} 0.7496 \tiny$\pm$0.0073} \\
CatBoost &
  {\color[HTML]{FF0000} 0.8057 \tiny$\pm$0.0046} &
  \textbf{0.6475 \tiny$\pm$0.0020} &
  0.5923 \tiny{$\pm$0.0268} &
  {\color[HTML]{3531FF} 0.5669 \tiny$\pm$0.0052} &
  0.9032 \tiny{$\pm$0.0007} &
  0.6861 \tiny{$\pm$0.0367} &
  \textbf{0.7506 \tiny$\pm$0.0144} \\ \hline
\end{tabular}}
    \caption{Comparison of F1 score between GFTab and baseline models on 21 tabular benchmark datasets in 10\% labeled training. The best performing method is highlighted in \textcolor{red}{red} and the second best in \textcolor{blue}{blue}, while the third best is \textbf{bold}.}
    \label{f1_01}
\end{table*}

\begin{table*}[!htbp]
\centering
    \centering
{\fontsize{6.7}{7.2}\selectfont
\begin{tabular}{lccccccc}
\hline
\multicolumn{8}{l}{\textbf{Panel A. Datasets with more categorical variables}} \\
\multicolumn{1}{c}{\textbf{model}} &
  \textbf{Diabetes} &
  \textbf{Insurance} &
  \textbf{adult} &
  \textbf{bank} &
  \textbf{cmc} &
  \textbf{credit-approval} &
  \textbf{credit-g} \\ \hline
GFTab &
  {\color[HTML]{FF0000} 0.3854 \tiny{$\pm$0.0154}} &
  {\color[HTML]{FF0000} 0.4915  \tiny$\pm$0.0028} &
  {\color[HTML]{FF0000} 0.7616 \tiny$\pm$0.0098} &
  {\color[HTML]{3531FF} 0.6761 \tiny$\pm$0.0093} &
  0.4238 \tiny$\pm$0.0219 &
  \textbf{0.7703 \tiny$\pm$0.0287} &
  0.4452 \tiny$\pm$0.0588 \\
GRANDE &
  {\color[HTML]{3531FF} 0.3771 \tiny$\pm$0.0054} &
  {\color[HTML]{3531FF} 0.4641 \tiny$\pm$0.0083} &
  \textbf{0.7208 \tiny$\pm$0.0049} &
  {\color[HTML]{FF0000} 0.6897 \tiny$\pm$0.0063} &
  0.3856 \tiny{$\pm$0.0109} &
  {\color[HTML]{3531FF} 0.8095 \tiny$\pm$0.0089} &
  {\color[HTML]{FF0000} 0.5809 \tiny$\pm$0.0464} \\
TabPFN &
  0.2339 \tiny{$\pm$0.0002} &
  0.4312 \tiny{$\pm$0.0012} &
  0.6950 \tiny{$\pm$0.0047} &
  0.5256 \tiny{$\pm$0.0032} &
  \textbf{0.4240 \tiny{$\pm$0.0041}} &
  {\color[HTML]{FF0000} 0.8463 \tiny$\pm$0.0093} &
  0.4118 \tiny{$\pm$0.0002} \\
SCARF &
  0.2350 \tiny{$\pm$0.0009} &
  0.4327 \tiny{$\pm$0.0022} &
  0.4443 \tiny{$\pm$0.0212} &
  0.4689 \tiny{$\pm$0.0110} &
  0.3438 \tiny{$\pm$0.0101} &
  0.6784 \tiny{$\pm$0.1083} &
  0.4421 \tiny{$\pm$0.0556} \\
SubTab &
  0.2211 \tiny{$\pm$0.0050} &
  \textbf{0.4372 \tiny{$\pm$0.0019}} &
  {\color[HTML]{3531FF} 0.7360 \tiny$\pm$0.0017} &
  0.6029 \tiny{$\pm$0.0231} &
  {\color[HTML]{3531FF} 0.4427 \tiny$\pm$0.0103} &
  0.5809 \tiny{$\pm$0.0190} &
  0.4925 \tiny{$\pm$0.0115} \\
VIME &
  0.2134 \tiny{$\pm$0.0097} &
  0.3921 \tiny{$\pm$0.0001} &
  0.6675 \tiny{$\pm$0.0082} &
  \textbf{0.6351 \tiny$\pm$0.0106} &
  {\color[HTML]{FF0000} 0.4749 \tiny$\pm$0.0691} &
  0.6440 \tiny{$\pm$0.1110} &
  {\color[HTML]{3531FF} 0.5464 \tiny{$\pm$0.0671}} \\
XGBoost &
  \textbf{0.3467 \tiny$\pm$0.0002} &
  0.4353 \tiny{$\pm$0.0019} &
  0.6840 \tiny{$\pm$0.0001} &
  0.5098 \tiny{$\pm$0.0014} &
  0.4079 \tiny{$\pm$0.0046} &
  0.7538 \tiny{$\pm$0.0113} &
  \textbf{0.5311 \tiny{$\pm$0.0214}} \\
CatBoost &
  0.3421 \tiny{$\pm$0.0040} &
  0.4312 \tiny{$\pm$0.0005} &
  0.6084 \tiny{$\pm$0.0006} &
  0.4696 \tiny{$\pm$0.0011} &
  0.3243 \tiny{$\pm$0.0052} &
  0.7406 \tiny{$\pm$0.0199} &
  0.4118 \tiny{$\pm$0.0005} \\ \hline
\multicolumn{1}{c}{\textbf{model}} &
  \textbf{dresses-sales} &
  \textbf{fars} &
  \textbf{jasmine} &
  \textbf{kick} &
  \textbf{okcupid-stem} &
  \textbf{online-shoppers} &
  \textbf{shrutime} \\ \hline
GFTab &
  0.4275 \tiny$\pm$0.1702 &
  {\color[HTML]{FF0000} 0.5333 \tiny$\pm$0.0079} &
  0.7010 \tiny{$\pm$0.0234} &
  {\color[HTML]{FF0000} 0.5201 \tiny$\pm$0.0055} &
  {\color[HTML]{3531FF} 0.3903 \tiny$\pm$0.0297} &
  \textbf{0.7090 \tiny{$\pm$0.0138}} &
  {\color[HTML]{3531FF} 0.6222 \tiny$\pm$0.0167} \\
GRANDE &
  {\color[HTML]{3531FF} 0.5633 \tiny{$\pm$0.0231}} &
  \textbf{0.5210 \tiny$\pm$0.0030} &
  {\color[HTML]{3531FF} 0.7461 \tiny$\pm$0.0097} &
  {\color[HTML]{3531FF} 0.5070 \tiny$\pm$0.0034} &
  {\color[HTML]{FF0000} 0.4114 \tiny$\pm$0.0031} &
  0.6163 \tiny{$\pm$0.0035} &
  0.5653 \tiny{$\pm$0.0085} \\
TabPFN &
  0.2891 \tiny{$\pm$0.0058} &
  0.4236 \tiny{$\pm$0.0012} &
  0.6811 \tiny{$\pm$0.0039} &
  0.4749 \tiny{$\pm$0.0201} &
  0.2995 \tiny{$\pm$0.0014} &
  0.6578 \tiny$\pm$0.0074 &
  0.5899 \tiny{$\pm$0.0126} \\
SCARF &
  0.5010 \tiny{$\pm$0.0284} &
  0.1238 \tiny{$\pm$0.0263} &
  0.5767 \tiny{$\pm$0.0285} &
  0.4775 \tiny{$\pm$0.0048} &
  0.2936 \tiny{$\pm$0.0061} &
  0.4732 \tiny{$\pm$0.0035} &
  0.4422 \tiny{$\pm$0.0011} \\
SubTab &
  {\color[HTML]{FF0000} 0.5770 \tiny$\pm$0.0190} &
  0.4386 \tiny{$\pm$0.0016} &
  0.5919 \tiny{$\pm$0.0066} &
  \textbf{0.4885 \tiny{$\pm$0.0007}} &
  \textbf{0.3862 \tiny{$\pm$0.0032}} &
  0.5975 \tiny{$\pm$0.0110} &
  {\color[HTML]{FF0000} 0.6313 \tiny$\pm$0.0026} \\
VIME &
  0.4380 \tiny{$\pm$0.1270} &
  {\color[HTML]{3531FF} 0.5331 \tiny$\pm$0.0148} &
  \textbf{0.7187 \tiny$\pm$0.0108} &
  0.4321 \tiny{$\pm$0.0015} &
  0.3083 \tiny{$\pm$0.0292} &
  0.4205 \tiny{$\pm$0.0033} &
  0.5400 \tiny{$\pm$0.0397} \\
XGBoost &
  0.4637 \tiny{$\pm$0.0002} &
  0.4379 \tiny{$\pm$0.0010} &
  0.6933 \tiny{$\pm$0.0067} &
  0.4749 \tiny{$\pm$0.0002} &
  0.3354 \tiny{$\pm$0.0020} &
  {\color[HTML]{FF0000} 0.7791 \tiny$\pm$0.0022} &
  \textbf{0.6026 \tiny$\pm$0.0222} \\
CatBoost &
  \textbf{0.5310 \tiny{$\pm$0.025}} &
  0.3568 \tiny{$\pm$0.0001} &
  {\color[HTML]{FF0000} 0.7675 \tiny$\pm$0.0014} &
  0.4749 \tiny{$\pm$0.0005} &
  0.3355 \tiny{$\pm$0.0170} &
  {\color[HTML]{3531FF} 0.7284 \tiny$\pm$0.0050} &
  0.4476 \tiny{$\pm$0.0038} \\ \hline
\multicolumn{8}{l}{} \\ \hline
\multicolumn{8}{l}{\textbf{Panel B. Datasets with more continouse variables}} \\
\multicolumn{1}{c}{\textbf{model}} &
  \textbf{KDD} &
  \textbf{Shipping} &
  \textbf{churn} &
  \textbf{eye-movements} &
  \textbf{nomao} &
  \textbf{qsar} &
  \textbf{road-safety} \\ \hline
GFTab &
  \textbf{0.7023 \tiny$\pm$0.0271} &
  0.6233 \tiny{$\pm$0.0090} &
  0.5641 \tiny{$\pm$0.0179} &
  0.5269 \tiny{$\pm$0.0107} &
  \textbf{0.8836 \tiny$\pm$0.0048} &
  {\color[HTML]{000000} \textbf{0.7389 \tiny$\pm$0.0198}} &
  0.7195 \tiny{$\pm$0.0091} \\
GRANDE &
  0.6810 \tiny{$\pm$0.0236} &
  0.6257 \tiny{$\pm$0.0041} &
  0.4806 \tiny{$\pm$0.0093} &
  0.5365 \tiny{$\pm$0.0046} &
  0.8764 \tiny{$\pm$0.0087} &
  0.7191 \tiny{$\pm$0.0277} &
  \textbf{0.7236 \tiny$\pm$0.0052} \\
TabPFN &
  0.6977 \tiny{$\pm$0.0012} &
  {\color[HTML]{3531FF} 0.6448 \tiny$\pm$0.0038} &
  0.4624 \tiny{$\pm$0.0020} &
  0.3336 \tiny{$\pm$0.0121} &
  0.7806 \tiny{$\pm$0.0223} &
  {\color[HTML]{3531FF} 0.7631 \tiny$\pm$0.0105} &
  0.6989 \tiny{$\pm$0.0062} \\
SCARF &
  0.5287 \tiny{$\pm$0.0259} &
  0.6061 \tiny{$\pm$0.0061} &
  0.4666 \tiny{$\pm$0.0083} &
  0.5020 \tiny{$\pm$0.0091} &
  0.5092 \tiny{$\pm$0.0106} &
  0.6427 \tiny{$\pm$0.0848} &
  0.5069 \tiny{$\pm$0.0092} \\
SubTab &
  0.5332 \tiny{$\pm$0.0077} &
  0.5451 \tiny{$\pm$0.0134} &
  \textbf{0.5752 \tiny$\pm$0.0141} &
  {\color[HTML]{FF0000} 0.5681 \tiny$\pm$0.0179} &
  0.8727 \tiny{$\pm$0.0055} &
  0.6689 \tiny{$\pm$0.0243} &
  0.6604 \tiny{$\pm$0.0025} \\
VIME &
  0.6530 \tiny{$\pm$0.0066} &
  0.5759 \tiny{$\pm$0.0411} &
  {\color[HTML]{FF0000} 0.6320 \tiny$\pm$0.0266} &
  0.5284 \tiny{$\pm$0.0308} &
  0.8536 \tiny{$\pm$0.0032} &
  {\color[HTML]{FF0000} 0.7631 \tiny$\pm$0.0044} &
  0.6853 \tiny{$\pm$0.0012} \\
XGBoost &
  {\color[HTML]{3531FF} 0.7618 \tiny$\pm$0.0218} &
  \textbf{0.6387 \tiny$\pm$0.0156} &
  0.5706 \tiny{$\pm$0.0044} &
  {\color[HTML]{3531FF} 0.5600 \tiny$\pm$0.0012} &
  {\color[HTML]{FF0000} 0.9030 \tiny$\pm$0.0019} &
  0.7314 \tiny$\pm$0.0148 &
  {\color[HTML]{FF0000} 0.7375 \tiny$\pm$0.0036} \\
CatBoost &
  {\color[HTML]{FF0000} 0.8188 \tiny$\pm$0.0078} &
  {\color[HTML]{FF0000} 0.6555 \tiny$\pm$0.0005} &
  {\color[HTML]{3531FF} 0.5896 \tiny$\pm$0.0093} &
  \textbf{0.5513 \tiny{$\pm$0.0022}} &
  {\color[HTML]{3531FF} 0.8918 \tiny$\pm$0.0004} &
  0.7192 \tiny{$\pm$0.0110} &
  {\color[HTML]{3531FF} 0.7360 \tiny$\pm$0.0061} \\ \hline
\end{tabular}}
    \caption{Comparison of F1 score between GFTab and baseline models on 21 tabular benchmark datasets in 10\% labeled training with 20\% label noise. The best performing method is highlighted in \textcolor{red}{red} and the second best in \textcolor{blue}{blue}, while the third best is \textbf{bold}.}
    \label{f1_11}
\end{table*}

\pagebreak
\newpage

\end{document}